\documentclass[10pt,twocolumn,letterpaper]{article}

\usepackage[pagenumbers]{cvpr} % To force page numbers, e.g. for an arXiv version

\usepackage{mathtools}
\usepackage{multirow}
\definecolor{cvprblue}{rgb}{0.21,0.49,0.74}
\usepackage[pagebackref,breaklinks,colorlinks,allcolors=cvprblue]{hyperref}

\def\paperID{3638} % *** Enter the Paper ID here
\def\confName{CVPR}
\def\confYear{2025}

\title{ArtWeaver: Advanced Dynamic Style Integration via Diffusion Model}

\author{Chengming Xu$^1$\quad  Kai Hu$^2$\quad  Qilin Wang$^3$\quad  Donghao Luo$^1$\quad  Jiangning Zhang$^1$\quad  Xiaobin Hu$^1$\\  Yanwei Fu$^3$\quad  Chengjie Wang$^1$ \\
  $^1$Youtu Lab, Tencent\quad $^2$Carnegie Mellon University\quad $^3$Fudan University
}

\begin{document}

\twocolumn[{%
    \renewcommand\twocolumn[1][]{#1}%
    \maketitle
    \begin{center}
	\centering
	\captionsetup{type=figure}
	\includegraphics[width=0.97\linewidth]{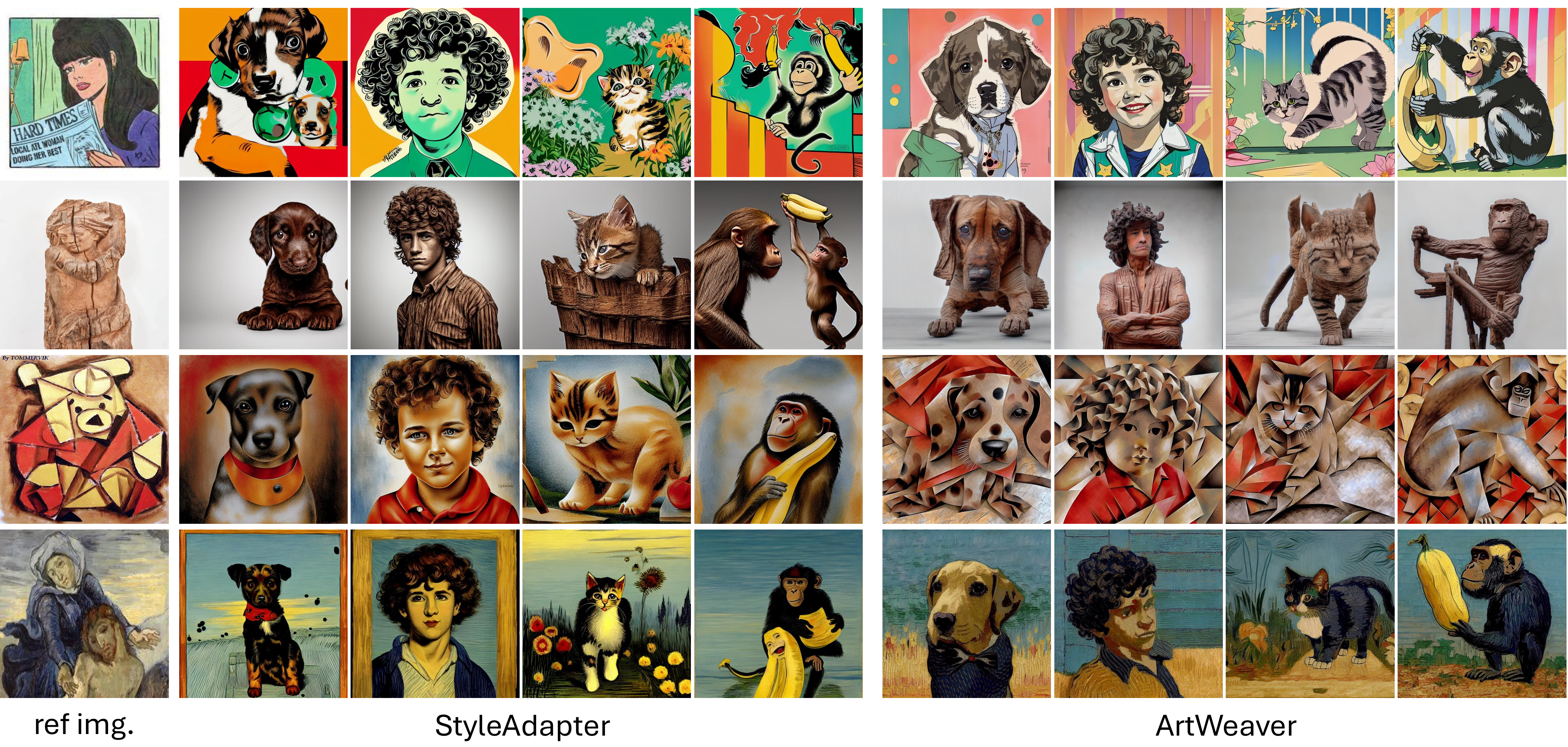} 
	\vspace{-4mm}
	\caption{Results generated by StyleAdapter and our proposed ArtWeaver. Our method can better stylized the images based on the reference images. Compared with StyleAdapter, our method can generate more consistent and delicate stylized images with different styles. For simplicity we only present the generated images here. For simplicity only one reference image for each style is shown. For the full reference images please refer to the supplementary material.}
	\label{fig:teaser}
    \end{center}%
}]
% \maketitle

% \begin{figure*}[t]
%     \centering
%     \includegraphics[width=0.9\textwidth]{figures/teaser_v2.pdf}
%     % \caption{Left: With our proposed ArtWeaver, the diffusion-based stylized image generation process can avoid issues such as inconsistent semantics and misinterpreted style, leading to better generation quality. Right: Our proposed method can generated consistent and delicate stylized images with different styles. For simplicity we only present the generated stylized images here. For the style reference images please refer to the supplementary material.}
%     \caption{Results generated by StyleAdapter and our proposed ArtWeaver. Our method can better stylized the images based on the reference images. Compared with StyleAdapter, our method can generate more consistent and delicate stylized images with different styles. For simplicity we only present the generated images here. For simplicity only one reference image for each style is shown. For the full reference images please refer to the supplementary material.}
%     \label{fig:teaser}
%     % \vspace{-0.1in}
% \end{figure*}

\begin{abstract}
    Stylized Text-to-Image Generation (STIG) aims to generate images from text prompts and style reference images. In this paper, we present ArtWeaver, a novel framework that leverages pretrained Stable Diffusion (SD) to address challenges such as misinterpreted styles and inconsistent semantics. Our approach introduces two innovative modules: the mixed style descriptor and the dynamic attention adapter. The mixed style descriptor enhances SD by combining content-aware and frequency-disentangled embeddings from CLIP with additional sources that capture global statistics and textual information, thus providing a richer blend of style-related and semantic-related knowledge. To achieve a better balance between adapter capacity and semantic control, the dynamic attention adapter is integrated into the diffusion UNet, dynamically calculating adaptation weights based on the style descriptors. Additionally, we introduce two objective functions to optimize the model alongside the denoising loss, further enhancing semantic and style consistency. Extensive experiments demonstrate the superiority of ArtWeaver over existing methods, producing images with diverse target styles while maintaining the semantic integrity of the text prompts.
    
\end{abstract}

\section{Introduction}

Stylized Image Generation (SIG), which involves generating images with specific artistic styles, has significant academic and practical implications, particularly in fields such as art and film. The emergence of diffusion-based generative models~\cite{ho2020ddpm, rombach2022ldm} has shifted the focus from traditional style transfer techniques to Stylized Text-to-Image Generation (STIG). In STIG, one or more style reference images are used as conditions to generate various images, incorporating additional information such as text prompts. Despite the complexity of handling mixed input conditions, STIG methods offer enhanced flexibility, making them highly applicable to real-world scenarios.

Recent advancements in Stable Diffusion (SD) based STIG methods, such as StyleAdapter~\cite{wangstyleadapter} and InstantStyle~\cite{wang2024instantstyle}, have demonstrated promising few-shot stylization capabilities. These methods extract style descriptors from reference images using a style embedding module, which are then injected into the diffusion UNet via cross-attention modules to guide the stylization process during denoising. This framework has been further extended in works like ArtAdapter~\cite{chen2024artadapter} and DEADiff~\cite{qi2024deadiff}. However, these methods still encounter significant challenges: (1) \textbf{misinterpreted style}, where the generated images do not fully capture the intricate styles of the reference images, and (2) \textbf{inconsistent semantics}, where elements from the reference images improperly influence the output, leading to misalignment with the text prompts.

These challenges primarily arise from the design of the style embedding extractor. Traditional methods often rely on a single-source style descriptor, using either pretrained models like CLIP~\cite{radford2021clip} or custom networks. However, reference images typically contain multi-level information, including local textures and global color schemes, along with rich semantic content which can bias the styles. Moreover, single-source style descriptors mix both low-frequency and high-frequency content, even though different denoising steps need different types of information. Consequently, such an approach can limit the style descriptor's ability to represent styles accurately, leading to less effective performance. Additionally, injecting style embeddings into all cross-attention layers of the diffusion UNet can bias the model's integration with text prompts, resulting in semantic inconsistencies.

In this paper, we introduce ArtWeaver, a novel STIG framework that features advanced techniques for \textit{extracting and injecting style information}. Our method builds upon the foundational pipeline of StyleAdapter, incorporating specific enhancements to improve style embedding extraction and injection. \textbf{For style extraction}, we propose a Mixed Style Descriptor (MSD). First, we follow StyleAdapter to adopt the CLIP-based patch descriptor, based on which we apply disentanglement according to the frequency domain. In addition, we incorporate a gram-based style descriptor via a Gram matrix~\cite{gatys2016nst} and a semantic descriptor derived from reference image captions. These diverse sources of information are integrated through transformer layers with adaptive scaling and shifting, where the semantic descriptor is aligned with and subtracted from the other descriptors. This approach captures a more comprehensive representation of target styles while preventing semantic leakage. \textbf{For style embedding injection}, a naive strategy as adopted by InstantStyle~\cite{wang2024instantstyle} is to limit the attachment of adapters to specific parts of the UNet, such as the upsampling layers, to prevent semantic distortion. However, this direct limitation can reduce the capacity of adapters and exacerbate the issue of misinterpreted style. To address this, we propose a Dynamic Attention Adapter (DAA) that generates sample-specific dynamic weights from the style embeddings. Compared with directly employing more layers for cross-attention modules, DAA leverages current intermediate features to adapt both self-attention and cross-attention layers in the diffusion UNet, allowing for more precise and flexible style adaptation. This ensures that the generated images maintain both the intended style and semantic consistency with the text prompts. 

Furthermore, we enhance the model with objectives beyond the standard noise prediction loss commonly used in diffusion models. We introduce a Gram consistency loss, augmenting the reference images with two sets of transformations: one set preserves the original style, while the other adopts a distorted style. We then compute Gram matrices of the estimated denoised results and these transformed reference images as their style-aware statistics. By applying a triplet loss among these matrices, the model is encouraged to generate images with more robust and consistent styles when processing different reference images. Additionally, we utilize a semantic disentanglement loss to mitigate the inconsistent semantics problem by contrasting the style embeddings against reference text embeddings, while ensuring they remain similar to the reference image embeddings.

% Furthermore, we enhance the model with objectives beyond the standard noise prediction loss commonly used in diffusion models. We introduce a Gram consistency loss, augmenting the reference images with two sets of transformations: one set preserves the original style, while the other adopts a distorted style. We then compute Gram matrices of the estimated denoised results and these transformed reference images as their style-aware statistics. By applying a triplet loss among these matrices, the model is encouraged to generate images with more robust and consistent styles when processing different reference images. Additionally, we utilize a semantic disentanglement loss to mitigate the inconsistent semantics problem by contrasting the style embeddings against reference text embeddings, while ensuring they remain similar to the reference image embeddings.

To demonstrate the effectiveness of our proposed method, we conduct extensive experiments across various styles, including both one-shot and multi-shot settings. Our results show that ArtWeaver significantly outperforms baseline methods like StyleAdapter, generating accurate styles and avoiding inconsistent semantics. In summary, the contributions of this work are as follows:
% \\
% 1) We present ArtWeaver, a novel framework for Stylized Text-to-Image Generation (STIG) that enhances the existing StyleAdapter pipeline. 
\\
1) Our ArtWeaver introduce the Mixed Style Descriptor (MSD) which captures comprehensive target style representations while preventing semantic leakage through a negative embedding branch.
% a multi-source approach that dynamically integrates CLIP-based patch descriptor, gram matrix-based descriptor, and semantic descriptor from image captions, capturing comprehensive target style representations while preventing semantic leakage through a negative embedding branch.
\\
2) To enhance style embedding injection, ArtWeaver introduces a dynamic attention adapter that generates weights from style embeddings, enabling precise adaptation of self-attention and cross-attention layers in the diffusion UNet. 
% This ensures that generated images maintain the intended style and semantic consistency with text prompts.
\\
3) Our model incorporates objectives beyond standard noise prediction loss, including a novel Gram consistency loss that promotes robust and consistent styles through triplet loss on Gram matrices from transformed reference images. Additionally, a semantic disentanglement loss contrasts style embeddings with reference text embeddings while maintaining similarity to reference image embeddings, addressing inconsistent semantics.

% Finally, extensive experiments demonstrate that ArtWeaver significantly outperforms baseline methods, including StyleAdapter, in generating correct styles and maintaining semantic consistency across various styles in both one-shot and multi-shot settings.

% \begin{itemize}
%     \item We propose a novel method for reference-based stylized image generation named ArtWeaver, boosting Stylized Text-to-Image Generation (STIG) with a multi-source style embedder and a dynamic attention adapter, greatly solving problems including misinterpreted style and inconsistent semantics.
%     \item Two objective functions are introduced to enhance our model, including a gram stylization loss and a semantic disentangle loss, which encourage the model to decouple the style related information from reference images with other style unrelated information.
%     \item Extensive experiment results show that our method consistently work well in both text-to-image and image-to-image tasks with different reference images, showing the practical value of such a method.

% \end{itemize}

\section{Related Works}
\noindent\textbf{Text-to-image diffusion models.} Diffusion models have been proven to be powerful generative models. DDPM~\cite{ho2020ddpm} originated to propose the framework by modeling the mapping between Gaussian distribution and image distribution with the forward diffusion and inverse denoising process. Based on that Latent Diffusion Model (LDM)~\cite{rombach2022ldm} largely improved the practical usage by leveraging diffusion model to latent space instead of pixel space, which leads to commonly-known text-to-image diffusion models such as Stable Diffusion (SD), Midjourney and DALLE-3~\cite{betker2023dalle3}. Other works focus on improve the diffusion model structure. For example, DiT~\cite{peebles2023dit}, MDT~\cite{gao2023mdt} and PIXART-$\alpha$~\cite{chen2023pixart} utilize the transformer instead of UNet structure, which can be better scaled to larger model size. \cite{blattmann2022retrieval} and \cite{zhang2023remodiffuse} leverage ideas of Retrieval Augmented Generation (RAG) to generate images based on other retrived images which provide extra knowledge. \cite{yang2024mastering} propose to leverage the LLMs for planning the text-to-image problems. Different from the previous works based on SD, we focus on designing extra attention adaptation so that the knowledge contained in the style reference images can be smoothly embedded into the denoising process, leading to stylized images.

\noindent\textbf{Stylized image generation.} Among all conditional image generation tasks, stylized image generation has long been a highlighted one. Most previous works focus on style transfer, \textit{i.e.}, transfer the style of a content image given another style reference image. For example, MicroAST~\cite{wang2023microast} proposed to speed up such framework by abandoning the complex visual encoder and utilizing a dual-modulation strategy. InST~\cite{zhang2023inst} realized style transfer by inverting the content image to noise and then re-generate it with the condition control of style images. StyleAdapter~\cite{wangstyleadapter} and ArtAdapter~\cite{chen2024artadapter} proposed a new framework which can generate images directly from style reference images and text prompts without content images. StyleID~\cite{chung2024styleid} adopted training-free approach to inverse both content image and style image into noises than merge them. DEADiff~\cite{qi2024deadiff} utilizes a pipeline similar to StyleAdapter, but leverages a self-constructed high-quality dataset. Somepalli et. al.~\cite{somepalli2024measuring} proposed a style descriptor with contrastive learning. StyleTokenizer~\cite{li2024styletokenizer} built a new tokenizer to align the latent space of style and textual information. Our work mainly follows StyleAdapter to present a generalized stylization method. Different from StyleAdapter, we analyze the role of style reference images and text prompts in the generation process. Based on that, we propose a novel module to extract more representative style embeddings, which are then injected into noise space with our proposed dynamic adapter.

\section{Preliminary: Stable Diffusion \label{sec:preliminary}}
 % Diffusion models are probabilistic models that learn a data distribution $p_\theta\left(\mathbf{x}_0\right)$ by progressively denoising a standard Gaussian distribution, which can be computed by iteratively adding Gaussian noise to the clean data $\mathbf{x}_0$. 

Diffusion models model the data distribution $p_\theta\left(\mathbf{x}_0\right)$ of clean data $\mathbf{x}_0$ by progressively denoising a standard Gaussian distribution, of which the learning process is instantiated as denoising score matching. 
 % as $p_\theta\left(\mathbf{x}_0\right)=$ $\int p_\theta\left(\mathbf{x}_{0: T}\right) d \mathbf{x}_{1: T}$, where $\mathbf{x}_{1: T}$ represents intermediate denoising results. The forward process of diffusion models is a Markov Chain that iteratively adds Gaussian noise to the clean data $\mathbf{x}_0$. 
 % , which maps data distribution to Guassian distribution
 Stable Diffusion (SD) extends such a model to text-to-image based on text prompt $p$. With pre-trained VQ-VAE~\cite{van2017vqvae} containing encoder $\mathcal{E}$ and decoder $\mathcal{D}$, SD allows the model to focus more on the semantic information of data and improves efficiency. A diffusion UNet is used to predict the noise, in which attention mechanism is adopted. Specifically, for the $l$-th layer, self-attention is first used to interact among spatial features: $z^l=Attention(W_Q^l  z^l, W_{K}^l  z^l, W_{V}^l  z^l)$, where $Attention$ denotes the attention operator, $z^l$ denotes latent embeddings of the $l$-th layer, $W_Q, W_K, W_V$ denotes the projection layers of self-attention. After that the cross-attention is utilized to merge condition information such as text prompt: $\hat{z}^l=Attn(\hat{W}_{Q_t}^l  z^l, \hat{W}_{K_t}^l  z_{text}, \hat{W}_{V_t}^l  z_{text})$, where $z_{text}$ denotes text prompt embedding, $\hat{W}_Q, \hat{W}_K, \hat{W}_V$ denotes the projection layers of cross-attention. The training objective of SD is as follows:
\begin{equation}
\label{eq:ldm}
\mathcal{L}_{noise}=\mathbb{E}_{\mathcal{E}(x), \epsilon \sim \mathcal{N}(0,1), t}\left[\left\|\epsilon-\epsilon_{\theta}\left(z^t, t\right)\right\|_2^2\right],
\end{equation}
where $t$ is uniformly sampled from $\{0,...,T\}$, $z^t$ denotes noisy latent at $t$-th timestep.

\section{Methodology}
\begin{figure*}
    \centering
    \includegraphics[width=0.9\textwidth]{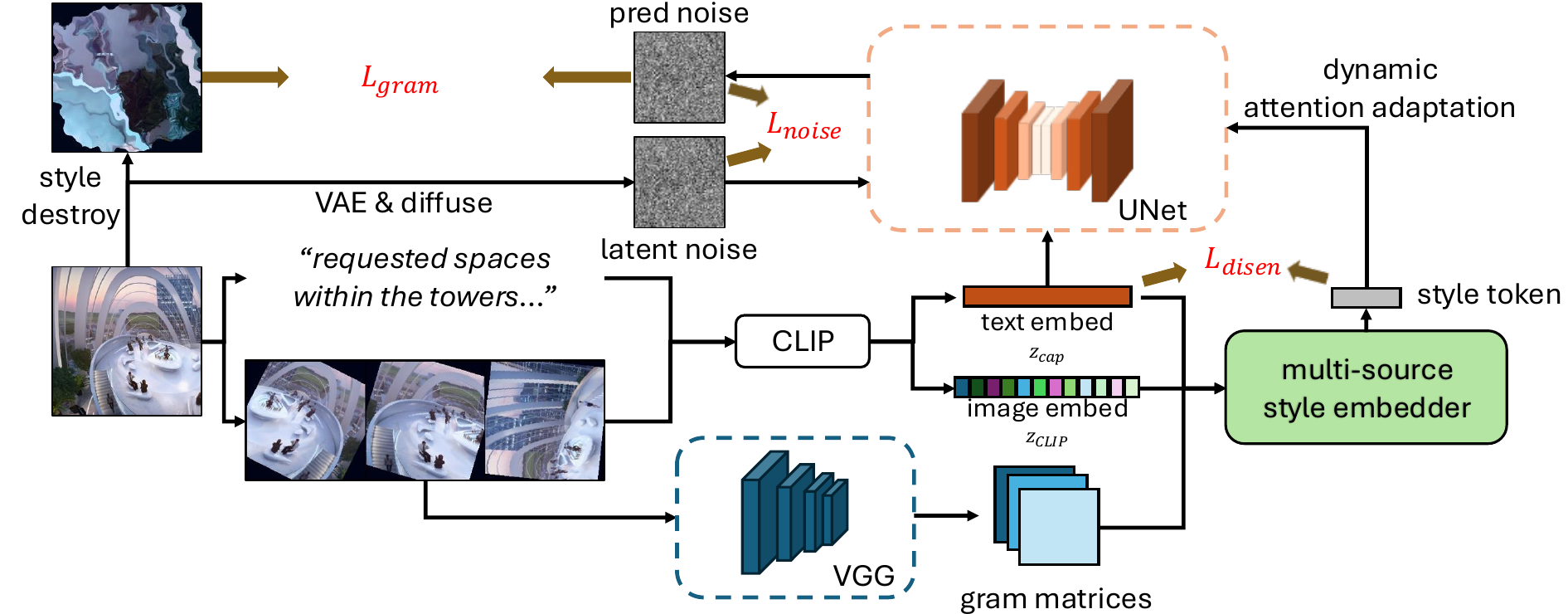}
    \caption{Overview of our proposed ArtWeaver. Concretely, the \textbf{Mixed Style Descriptor (MSD)} (Sec.~\ref{sec:style}) aggregates style patterns contained in each reference image simultaneously from global and local levels. Style embeddings are then engaged in the denoising procedure through the proposed \textbf{Dynamic Attention Adaptation (DAA)} (Sec.~\ref{sec:attention}), which guides both the attentions in diffusion UNet to properly merge style and semantic information from different sources. During training augmented input images are used as style reference images, through which objectives as described in Sec.~\ref{sec:loss} are used to optimize the model. During inference the style reference images are achieved with manual assignment instead of using augmentation of a specific image.}
    \label{fig:model}
\end{figure*}

We focus on reference-based Stylized Text-to-Image Generation (STIG) in this paper. Formally, a reference style image set $\mathcal{I}_s=\{\mathbf{I}_{style}^i\}_{i=1}^{N_s}$, where $N_s$ denotes the number of reference images, together with text prompt $p$ are given as condition information. $N_s$ can be variable among different trials to describe different style concepts. The model is required to generate image $\mathbf{I}$ that shares the same style pattern with $\mathcal{I}_s$ and same semantic meaning with text prompt $p$. To solve this task we present a novel framework named ArtWeaver based on SD, as shown in Fig.~\ref{fig:model}, which will be introduced in this section.

\begin{figure}
    \centering
    \includegraphics[width=0.9\linewidth]{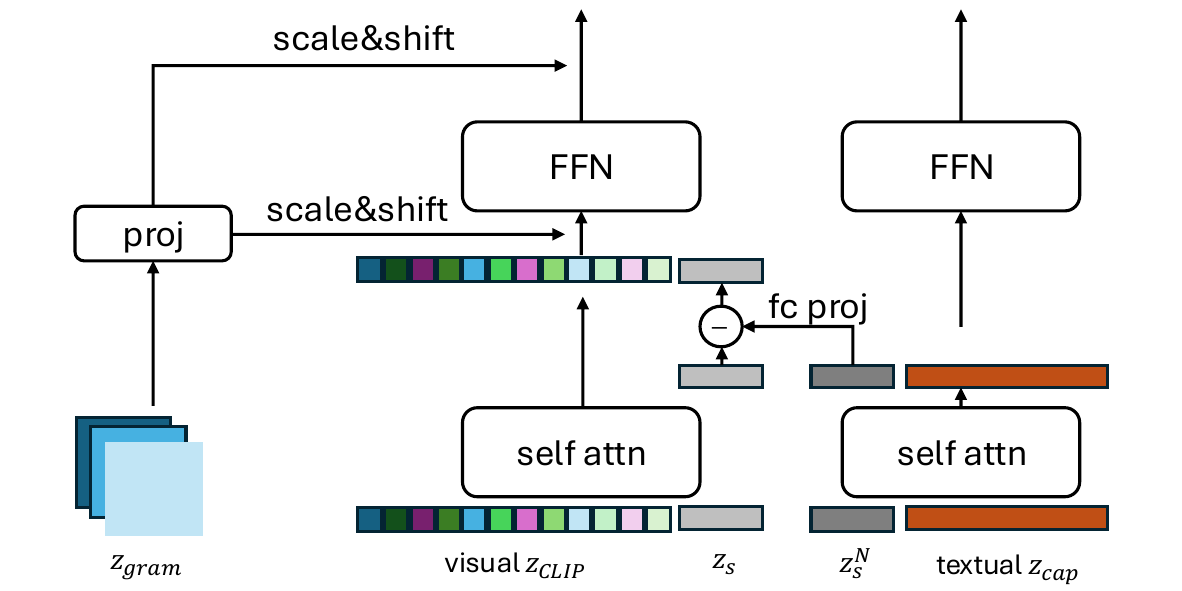}
    \caption{The structure of the Mixed Style Descriptor (MSD)}
    \label{fig:MSD}
    \vspace{-0.2in}
\end{figure}

\subsection{Mixed Style Descriptor \label{sec:style}}

% Given the successful application of text-to-image based on SD, an ideal reference-based stylization should necessarily depend on extracting style embeddings that are as representative as the text embeddings resulted from strong embedders such as CLIP and T5, from reference images. StyleAdapter~\cite{wangstyleadapter} adopts CLIP visual encoder~\cite{radford2021clip} to extract the patch-level features from style reference images, which are then processed with cross-attentions using learnable style tokens. However, StyleAdapter still suffers from issues such as misinterpreted style and inconsistent semantics, \textit{i.e.}, the semantic knowledge in reference images is generated during the denoising procedure. The former problem can be attributed to the fact that CLIP-based patch embeddings concentrate more on local patterns. In this way, the style embeddings fail to hold the global style patterns, leading to misinterpreted styles. Meanwhile, since CLIP image embeddings are aligned with their captions, they inevitably contain information about content in the images. Incorporating such information in style embeddings can lead to inconsistent semantics. Moreover, since SD models different kinds of information across different denoising timesteps, providing identical conditional information means SD has to extract the required information on its own, thus leading to heavier learning burden for SD. 

Given the successful application of text-to-image generation based on Stable Diffusion (SD), an ideal reference-based stylization should rely on extracting style embeddings that are as representative as the text embeddings produced by representative models like CLIP and T5. However, previous methods still face issues such as misinterpreted styles and inconsistent semantics. These issues arise because the previously adopted style descriptor is limited to both comprehensively represent the style and get rid of the negative effect of semantic information. Moreover, as SD models different kinds of information across various denoising timesteps, providing identical and entangled frequency-related information forces SD to extract the required information independently, increasing its learning burden.

To address these problems, we introduce a comprehensive Mixed Style Descriptor (MSD). Specifically, given a style image set $\mathcal{I}_s$, we extract three different types of features using pretrained models: (1) \textbf{CLIP-based style descriptor}: Following StyleAdapter, we use CLIP to encode each style image $\mathbf{I}_{style}^i$ into latent patch tokens $\tilde{z}_{CLIP}^i\in\mathbb{R}^{c\times (wh)}$, where $c$, $h$, and $w$ denote the latent channel, height, and width of CLIP features. We then apply discrete wavelet transform (DWT) to $\tilde{z}_{CLIP}^i$, resulting in low-frequency features $\tilde{z}_{CLIP, lf}^i$ and high-frequency counterparts $\tilde{z}_{CLIP, hf}^i$. All low and high-frequency features from $\mathcal{I}_s$ are concatenated along the token dimension, forming $z_{CLIP}$. (2) \textbf{Gram-based style descriptor}: Inspired by Neural Style Transfer (NST)~\cite{gatys2016nst}, we use the Gram matrix to complement the style information. Specifically, following NST, $\mathbf{I}_{style}^i$ is processed with pretrained VGG-19~\cite{simonyan2014vgg} to obtain the \textit{relu3\_1} feature $z_{vgg}^i\in \mathbb{R}^{c'\times h'w'}$, where $c'$, $h'$, and $w'$ denote the latent channel, height, and width of the VGG feature map. The Gram matrix is then calculated as $z_{gram}^i=z_{vgg}^i  {z_{vgg}^i}^T$, which is flattened to a vector with dimension $\mathbb{R}^{c^2}$. The Gram matrices for different reference images are averaged to obtain the final representation $z_{gram}$. (3) \textbf{Semantic descriptor}: We use the CLIP text encoder to extract the text embedding $z_{cap}^i$ of the captions of $\mathbf{I}_{style}^i$, which are then concatenated into $z_{cap}$. During training, the captions are provided in the training set. During inference, we use BLIP~\cite{li2022blip} to annotate the caption of style reference images, which can be replaced with other advanced image caption models for future works.

Generally, each token in $z_{CLIP}$ contains style features related with image content and disentangled in the frequency domain, while $z_{gram}$ focuses less on image content but more on global statistics related with the style of interest. On the other hand, $z_{cap}$ contains the semantic information of $\mathcal{I}_s$, which should be eliminated in the final style descriptor to avoid inconsistent semantics. To properly make use of these descriptors, we propose a novel dual-branch structure as shown in Fig.~\ref{fig:MSD}. Concretely, several learnable style tokens $z_s$ are first attached to $z_{CLIP}$, with their replication $z^N_s$ denoted as negative semantic tokens attached to $z_{cap}$:
\begin{equation}
    \hat{z}_{CLIP} = z_{CLIP} \|_t (z_s+\delta_t),\quad    \hat{z}_{caption} = z_{cap} \|_t z^N_s
\end{equation}
where $ \|_t $ denotes concatenation along the token dimension, $\delta_t$ is the same time embedding of the denoising timestep as used in diffusion UNet. $\hat{z}_{caption}$ is then individually processed with several transformer layers to aggregate the information between style tokens and text embedding. For $\hat{z}_{CLIP}$, we adopt a modified version of transformer layer. Specifically, $z_{gram}$ is first projected to scaling and shift coefficients. These coefficients are applied to the self-attention procedure the same as adaLN~\cite{perez2018film}. Before processing the attention result with FFN, the style token part in $\hat{z}_{caption}$ is projected with MLP to align with the image latent space and subtracted from its counterpart in  $\hat{z}_{CLIP}$.

Our design offers three major advantages. First, the module is timestep-aware. Since the frequency-aware style tokens are merged with the time embedding, different denoising steps can model different information. Second, $z_{gram}$ can well complement the information provided by $z_{CLIP}$ to better focus on style-related knowledge, thus avoiding irrelevant but repetitive local contents shared among style images. Third, the negative semantic tokens generally contain more abstract content information rather than style. Consequently, subtracting them from $\hat{z}_{CLIP}$ can help alleviate the problem of inconsistent semantics. While some captions may describe the style of images, the model can learn to maintain this knowledge during training thanks to the training strategy described in Sec.~\ref{sec:loss}. By using the proposed module, we can learn more representative and generalizable style embeddings, which can better facilitate the stylization process described as follows.

\begin{figure}
    \centering
    \includegraphics[width=0.95\linewidth]{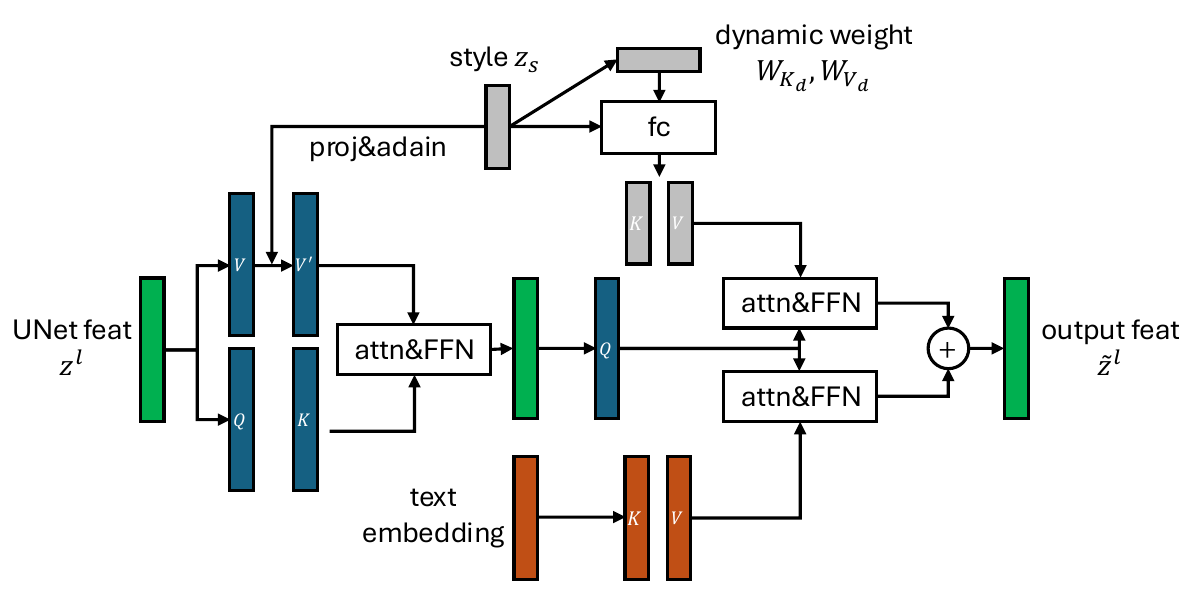}
    \caption{The structure of the Dynamic Attention Adapter (DAA)}
    \label{fig:DAA}
    \vspace{-0.2in}
\end{figure}

\subsection{Dynamic Attention Adaptation \label{sec:attention}}

The extracted style embedding $z_s$ from $\mathcal{I}_s$ as described above can then be used to adapt the pretrained SD to guide the denoising process based on style information. Thanks to the design of self-attention and cross-attention mechanism in diffusion UNet, such adaptation can be simply instantiated as an extra cross-attention module that is parallel to the original prompt-based cross-attention, as adopted in StyleAdapter. However, we empirically find that such a method is suboptimal in a large amount of cases, leading to severe semantic inconsistency. A straightforward solution is cut down the number of extra attention modules so that only upsample layers in diffusion UNet are adapted, as in DEADiff and InstantStyle. In this way, the text prompt can dominate the cross-attention in half of the UNet, consequently resulting in better semantics in the generated images. However, this can decrease the capacity of adapters, leading to less preferable stylization. To this end, we propose adopting a Dynamic Attention Adaptation (DAA) strategy which is applied to both self-attention and cross-attention (Fig.~\ref{fig:DAA}). 

\noindent\textbf{Dynamic self-attention adapter.} As discussed in previous works~\cite{hertz2023stylealign}, the projected value tensor $W_V^l  z^l$ in the self-attentions contributes to the texture of generated images. Therefore we introduce a dynamic self-attention adapter module based on adaIN. Formally, for the $l$-th self-attention layer, we project $z_s$ with a linear layer and adjust $W_V^l  z^l$ according to statistics of $z_s$:
\begin{equation}
    \hat{\mathbf{V}}^l = \mu(f_{SA}^l(z_s)) + \frac{\sigma(f_{SA}^l(z_s))}{\sigma(W_V^l  z^l)}(W_V^l  z^l-\mu(W_V^l  z^l))
\end{equation}
where $f_{SA}^l$ denotes dynamic projection layer, $\mu, \sigma$ denote mean and standard deviation. By rescaling $\mathbf{V}^l$, the information contained in $z_s$ can be directly embedded into the image feature without destroying the structure and semantic meaning of the generated image.

\noindent\textbf{Dynamic cross-attention adapter.} To adapt the cross-attention layers, we follow the idea of StyleAdapter to adopt the dual-path cross-attention mechanism. Basically, for the $l$-th cross-attention layer, besides the original cross-attention performed between text embedding $z_{text}$ and image embedding $z^l$ as in Sec.~\ref{sec:preliminary}, an extra style-aware cross-attention is added as   $\tilde{z}^l=Attn(\hat{W}_{Q_t}^l  z^l, \hat{W}_{K_s}^l  z_s, \hat{W}_{V_s}^l  z_s)$
with additional learnable parameters $\hat{W}_{K_s}^l,\hat{W}_{V_s}^l$. Then $\hat{z}^l+\lambda \tilde{z}^l$ is fed into the following feed-forward networks, where $\lambda$ is learnable coefficient.

To enhance the capacity, we further propose a dynamic cross-attention adapter. Specifically, we first project the statistics of $z_s$ to a layer specific latent space:
\begin{equation}
    z_s^l = f_{proj-CA}^l(\mu(z_s)\|_c\sigma(z_s))
\end{equation}
where $f_{proj-CA}$ denotes a linear projection layer, $ \|_c $ denotes concatenation along the channel dimension. Then two weight generators instantiated as linear layers are applied to $z_s^l$, resulted in two dynamic weights $W_{K_d}^l, W_{V_d}^l$ with dimension $d_{}*d^l$. These two features are reshaped into linear layer weights and used to transform $z_s$. After that the style-aware cross-attention is modified as 
 \begin{equation}
     \tilde{z}^l=Attn(\hat{W}_{Q_t}^l  z^l, (\hat{W}_{K_s}^l+W_{K_d}^l)  z_s, (\hat{W}_{V_s}^l+W_{V_d}^l)  z_s)
 \end{equation}
 In this way, the key and value projections are partially dependent on $z_s$, resulting in a more complex transformation of $z_s$ and leading to better capacity. To make this module parameter-efficient, we adopt a grouping strategy, \textit{i.e.}, channels of $z_s$ in each group share the same dynamic weight to produce $W_{K_d}^l$ and $W_{V_d}^l$, hence lighter weight generators can be used to generate dynamic weights.

\subsection{Training Objectives \label{sec:loss}}
To train the model such that it can generate images that are conforms to both the style information from style reference images and semantic information from text prompts, we introduce a mixed training objectives including three terms as follows.
\begin{equation}
    \mathcal{L} = \mathcal{L}_{noise} + \mathcal{L}_{disen} + \mathcal{L}_{style}
    \label{eq:loss}
\end{equation}
where $\mathcal{L}_{noise}$ is the noise prediction loss as in Eq.~\ref{eq:ldm}. $\mathcal{L}_{disen}$ denotes a semantic disentangle loss applied to style embedding $z_s$. To ensure that the proposed style embedding model can get rid of the semantic information contained in $\mathbf{I}_{style}$ when producing $z_s$, $\mathcal{L}_{disen}$ is designed by enlarging the similarity between $z_s$ and $z_{CLIP}$ while decreasing the similarity between $z_s$ and text embedding $z_{text}$. Formally, $L_{disen} = sim(z_{cap}, z_s)-\delta sim(z_{CLIP}, z_s)$, where $\delta$ is hyper-parameter set as 0.1, $sim$ denotes cosine similarity. As discussed in Sec.~\ref{sec:style}, the text embedding represents more abstract semantic information than the image embedding. Therefore this loss term can help the model  avoid the possibility of semantic leakage, thus leading to better style embedding. On the other hand, to enhance the style consistency, we propose to regulate the Gram matrix of $\hat{x}^0$, which is the noisy estimation from $z^t$ and can be calculated as
\begin{equation}
    \hat{z}^0=\frac{z^t-\sqrt{1-\bar{\alpha}^t}\epsilon^t}{\sqrt{\bar{\alpha}^t}}, \quad \hat{x}^0=\mathcal{D}(\hat{z}^0)
\end{equation}
Specifically, we apply several rigid transformations such as random rotation and cropping to $\mathcal{I}_S$ to get a new image $\mathbf{I}_{pos}$ that has the same style as $\hat{x}_0$. Then elastic transformation and color jitter are also applied as \textit{style destroy} method to $\mathcal{I}_S$. The resulted $\mathbf{I}_{neg}$, while sharing similar semantic object to $\mathbf{I}_{inp}$, barely inherits the style from it. Then the objective is a triplet loss which can be written as
\begin{align}
    \delta_{p} &= \sum \left|\mathcal{G}(\phi_{vgg}(\hat{x}_0)) - \mathcal{G}(\phi_{vgg}(\mathbf{I}_{pos}))\right| \\
    \delta_{n} &= \sum \left|\mathcal{G}(\phi_{vgg}(\hat{x}_0)) - \mathcal{G}(\phi_{vgg}(\mathbf{I}_{neg}))\right| \\
    \mathcal{L}_{style} &= max \{\delta_{p}-\delta_{n}+0.1, 0\}
\end{align}
where $\mathcal{G}$ denotes the Gram matrix of features. By optimizing this loss term, the model is encouraged to learn more detailed style information, thus leading to better results. In total, during training, only the style embedder and the added adapters are trained with objectives as in Eq.~\ref{eq:loss}. Those used backbones such as VGG, CLIP and original parameters from SD are not trained. During inference, we directly use the trained models to generate stylized images without any test-time optimization.

\section{Experiments \label{sec:experiment}}
\subsection{Implementation Detail \label{sec:experiment-detail}}
\noindent\textbf{Dataset.} We follow StyleAdapter to adopt LAION-Aesthetic 6.5+ as the training set, which contains about 600k images. For each input image during training, we use its augmented variants as the style reference images. For evaluation we adopt 50 prompts used in StyleAdapter, and select 20 styles covering color, texture and global layout. More details are presented in the supplementary material. 

\noindent\textbf{Experiment setting.} Our experiments cover both one-shot and multi-shot settings. To make the evaluation more challenging, the number of reference images varies from 2 to 5 among different styles in multi-shot setting. We use all 20 styles for multi-shot experiments and 10 of them for one-shot experiments. Our proposed method does not need test-time optimization for both settings. In multi-shot experiment, the style descriptors produced by the MSD for all reference images are concatenated as the input condition.

\noindent\textbf{Training details.} We adopt AdamW as optimizer with 1e-5 learning rate. Our model is trained for 200,000 iterations on 8 V100s with 8 batch size on each gpu, which takes about 3 days. Our code will be release.

\noindent\textbf{Competitor.} We include extensive methods as our competitor. For 1-shot experiment, MicroAST (MAST)~\cite{wang2023microast}, StyleAdapter (SAda)~\cite{wangstyleadapter}, StyleDrop (SDrop)~\cite{sohn2023styledrop}, InstantStyle (InsStyle)~\cite{wang2024instantstyle}, DEADiff~\cite{qi2024deadiff} and StyleID~\cite{chung2024styleid} are adopted. For multi-shot experiment, InST~\cite{zhang2023inst}, LoRA~\cite{hu2021lora}, Textual Inversion (TI)~\cite{gal2022ti}, StyleDrop~\cite{sohn2023styledrop}, StyleAdapter and InstantStyle are adopted. Among all methods, MicroAST, InST, LoRA, and TI require test-time optimization, and the others are finetune-free methods.

\begin{table}[htb]
\parbox{.45\linewidth}{
\caption{Quantitative results for one-shot setting.}
\vspace{-0.1in}
\centering
\resizebox{\linewidth}{!}{%
\begin{tabular}{cccc}
\toprule
Backbone & 1-shot Methods & Text Sim $\uparrow$ & Style Sim $\uparrow$  \tabularnewline
\midrule
\multirow{6}{*}{SD15} & MicroAST~\cite{wang2023microast} & 0.299 & 0.529 \tabularnewline
& StyleDrop~\cite{sohn2023styledrop} & 0.290 & 0.583 \tabularnewline
& InstantStyle~\cite{wang2024instantstyle} & 0.167 & \textbf{0.840} \tabularnewline
& StyleAdapter~\cite{wangstyleadapter} & 0.282 & 0.668 \tabularnewline
& DEADiff~\cite{qi2024deadiff} & 0.293 & 0.590 \tabularnewline
& StyleID~\cite{chung2024styleid} & 0.298 & 0.543 \tabularnewline
& Ours & \textbf{0.299} & 0.708 \\
\midrule
\multirow{3}{*}{SDXL} & StyleAlign~\cite{hertz2023stylealign} & 0.276 & 0.645 \tabularnewline
& InstantStyle~\cite{wang2024instantstyle} & 0.295 & 0.652 \tabularnewline
& Ours & \textbf{0.311} & \textbf{0.696} \\
\bottomrule
\end{tabular}
}
\label{tab:1-shot}
}
\hfill
\parbox{.45\linewidth}{
\caption{Quantitative results for multi-shot setting.}
% \vspace{-0.1in}
\centering
\resizebox{\linewidth}{!}{%
\begin{tabular}{cccc}
\toprule
Backbone & multi-shot Methods & Text Sim $\uparrow$ & Style Sim $\uparrow$ \\
\midrule
\multirow{7}{*}{SD15} & InST~\cite{zhang2023inst} & 0.196 & 0.692 \tabularnewline
& LoRA~\cite{hu2021lora} & 0.237 & 0.665 \tabularnewline
& TI~\cite{gal2022ti} & 0.268 & 0.678 \tabularnewline
& StyleDrop~\cite{sohn2023styledrop} & 0.273 & 0.599 \tabularnewline
& InstantStyle~\cite{wang2024instantstyle} & 0.186 & \textbf{0.749} \tabularnewline
& StyleAdapter~\cite{wangstyleadapter} & 0.286 & 0.682 \tabularnewline
& Ours & \textbf{0.291} & 0.719 \\
\midrule
\multirow{2}{*}{SDXL} & InstantStyle~\cite{wang2024instantstyle} & 0.291 & 0.645 \tabularnewline
& Ours & \textbf{0.293} & \textbf{0.667} \\
\bottomrule
\end{tabular}
}
\label{tab:multi-shot}
}
\vspace{-0.2in}
\end{table}

\begin{figure*}[t]
    \centering
    \includegraphics[width=0.9\textwidth]{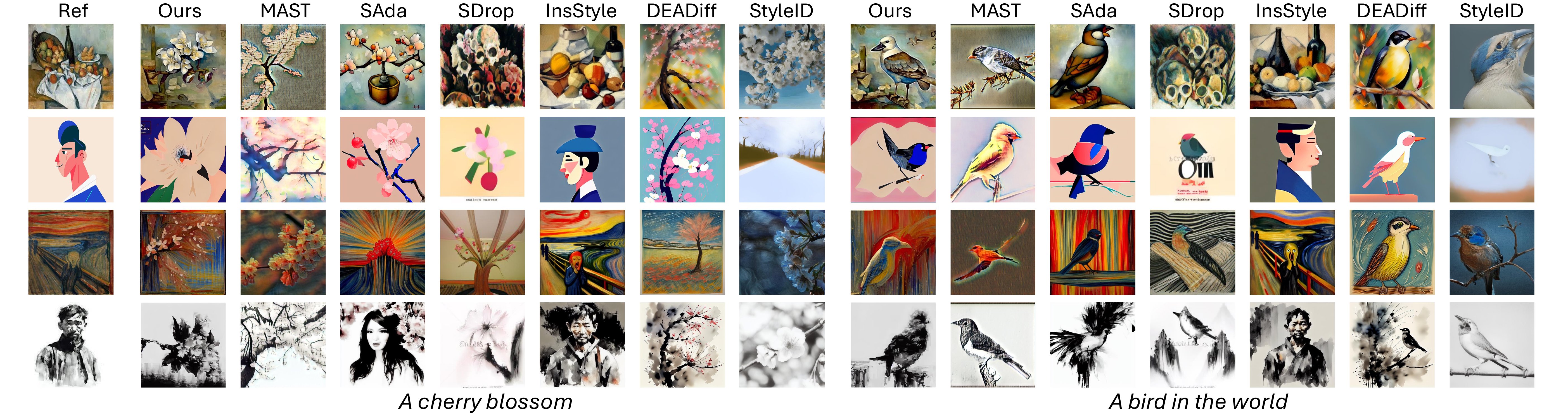}
    \caption{One-shot qualitative comparison with SD1.5 as backbone. For comparison with SDXL as backbone, please refer to the supplementary material. Zoom in for more details.}
    \label{fig:1-shot}
\end{figure*}

\begin{figure*}[htb]
    \centering
    \includegraphics[width=0.9\textwidth]{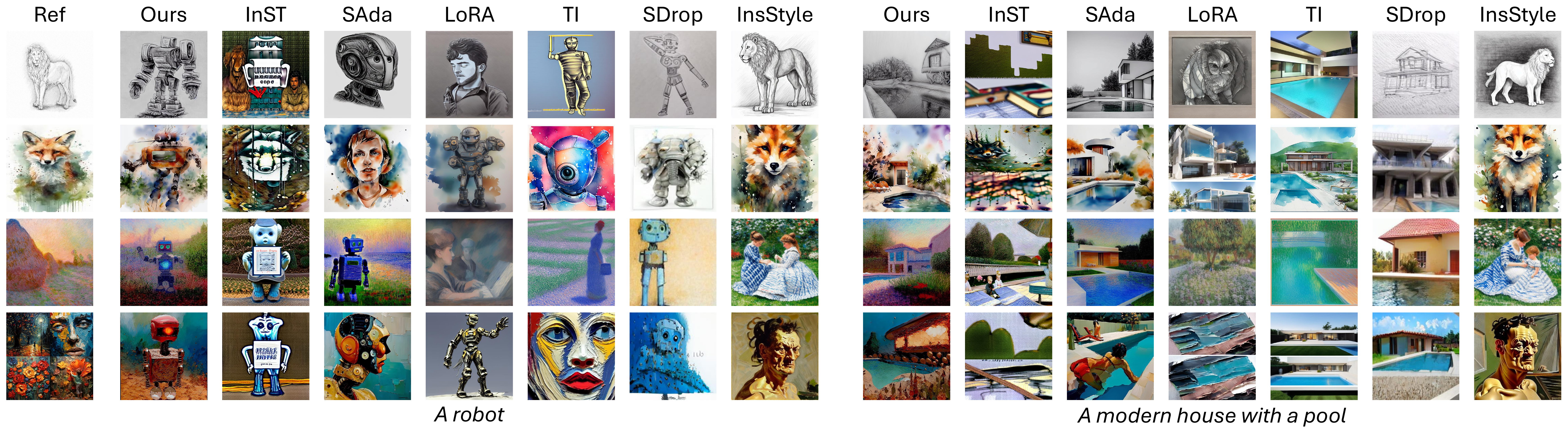}
    \vspace{-0.1in}
    \caption{Multi-shot qualitative comparison with SD1.5 as backbone. For detailed reference images and comparison with SDXL as backbone, please refer to the  supplementary material. Zoom in for more details.}
    \label{fig:multi-shot}
    \vspace{-0.2in}
\end{figure*}
 
\subsection{Quantitative Results}
% \noindent\textbf{Objective quantitative results.} 
For quantitative evaluation we adopt CLIP to calculate the style similarity between generated images and target style reference images, and the semantic similarity between generated images and target text prompts. The results are presented in Tab.~\ref{tab:1-shot} and Tab.~\ref{tab:multi-shot} for 1-shot and multi-shot respectively. Note that for SD1.5, InstantStyle receives incredibly high style similarity, which is not attributed to its strong performance. In fact, as we will show in the qualitative results, InstantStyle-SD1.5 totally leaks the content in the reference images into the generated images, thus leading to extremely poor text similarity. Moreover, MicroAST shares similar text similarity to ours one-shot experiments. This is because we use pretrained SD to generate base images for them, thus the basic semantic meaning is contained in the image. However the gap in terms of style similarity is much more marginal. As for SDXL, our method performs generally better than other competitors. The results for multi-shot setting are consistent with one-shot, thus showing the superiority of the proposed method.

\subsection{Qualitative Results}

We present several uncurated results with SD1.5 as backbone for both settings in Fig.~\ref{fig:1-shot} and Fig.~\ref{fig:multi-shot} respectively. First of all, as mentioned above, we find that InstantStyle suffers from problem of content leakage, resulting in meaningless generation. For one-shot experiment, since we use pretrained SD to first generate the content images, the style transfer based methods such as MicroAST can generally enjoy reasonable semantic consistency. However, such methods can only inherit the basic color information from reference images rather than the detailed style information such as shape, texture and layout, thus making them less preferable. For example, they fail to present the textures and curves. This is because these methods rely on simple representation to transfer the style-related knowledge from reference images, which leads to the problem of under-stylization. DEADiff, on the other hand, can stylize the images better. However, we find that the generated images of DEADiff, while being visually approvable, cannot follow the target style as shown in the reference images. Moreover, images generated by DEADiff seems to rely more on its learned prior knowledge rather than the input condition.
The performance of StyleAdapter is generally reasonable, while it is hard for this method to understand complex style patterns, leading to undesirable results when it comes to ink painting. Compared with these methods, our method can learn appropriate style information from reference images, \textit{e.g.}, the scattered color patches in the first and fourth row, and simultaneously keep the images faithful to the prompts, thus making the best of both worlds.

The multi-shot setting which is more challenging shows similar results. InST can hardly replicate the style. LoRA and TI suffer from limited style information. StyleAdapter, while utilizing a specifically-designed pipeline, shows a tendency to confuse the given styles with the photographic prior knowledge from pretrained SD. Such phenomenon is most obvious in the first row of Fig.~\ref{fig:multi-shot}, where pencil drawings are provided as reference images, but StyleAdapter generates grayscale photos. Our method, thanks to the proposed MSD which can extract more detailed style information and the dynamic attention adaptation, can generally generate different kinds of styles with high image quality and semantic fidelity.

\begin{table}[htb]
\caption{Ablation study for both one-shot and multi-shot settings.}
\vspace{-0.1in}
\label{tab:ablation-quant}
\centering
\resizebox{0.9\linewidth}{!}{
\begin{tabular}{ccccc}
\toprule
\multirow{2}{*}{\textbf{Methods}} & \multicolumn{2}{c}{One-shot} & \multicolumn{2}{c}{Multi-shot} \\
 & Text Sim & Style Sim & Text Sim & Style Sim  \tabularnewline
    \midrule
w/o Gram & 0.288 & 0.692 & 0.288 & 0.698 \tabularnewline
w/o NegEmb & 0.285 & 0.689 & 0.281 & 0.696 \tabularnewline
w/o OnlyUP & 0.286 & 0.697 & 0.286 & 0.702 \tabularnewline
w/o DA & \textbf{0.301} & 0.611 & \textbf{0.299} & 0.643 \tabularnewline
w/o $\mathcal{L}_{style}$ & 0.280 & 0.694 & 0.290 & 0.705 \tabularnewline
w/o $\mathcal{L}_{disen}$ & 0.282 & 0.695 & 0.286 & 0.694 \tabularnewline
Ours & 0.299 & \textbf{0.708} & 0.291 & \textbf{0.719} \\
\bottomrule
\end{tabular}
\vspace{-0.2in}
}
\end{table}

\subsection{Ablation Study}

To further verify the efficacy of our contributions, we conduct several ablation studies on multi-shot setting. The quantitative results are shown in Tab.~\ref{tab:ablation-quant}. More qualitative ablation studies are provided in the  supplementary material.

\begin{figure}[htb]
    \centering
    \includegraphics[width=0.95\linewidth]{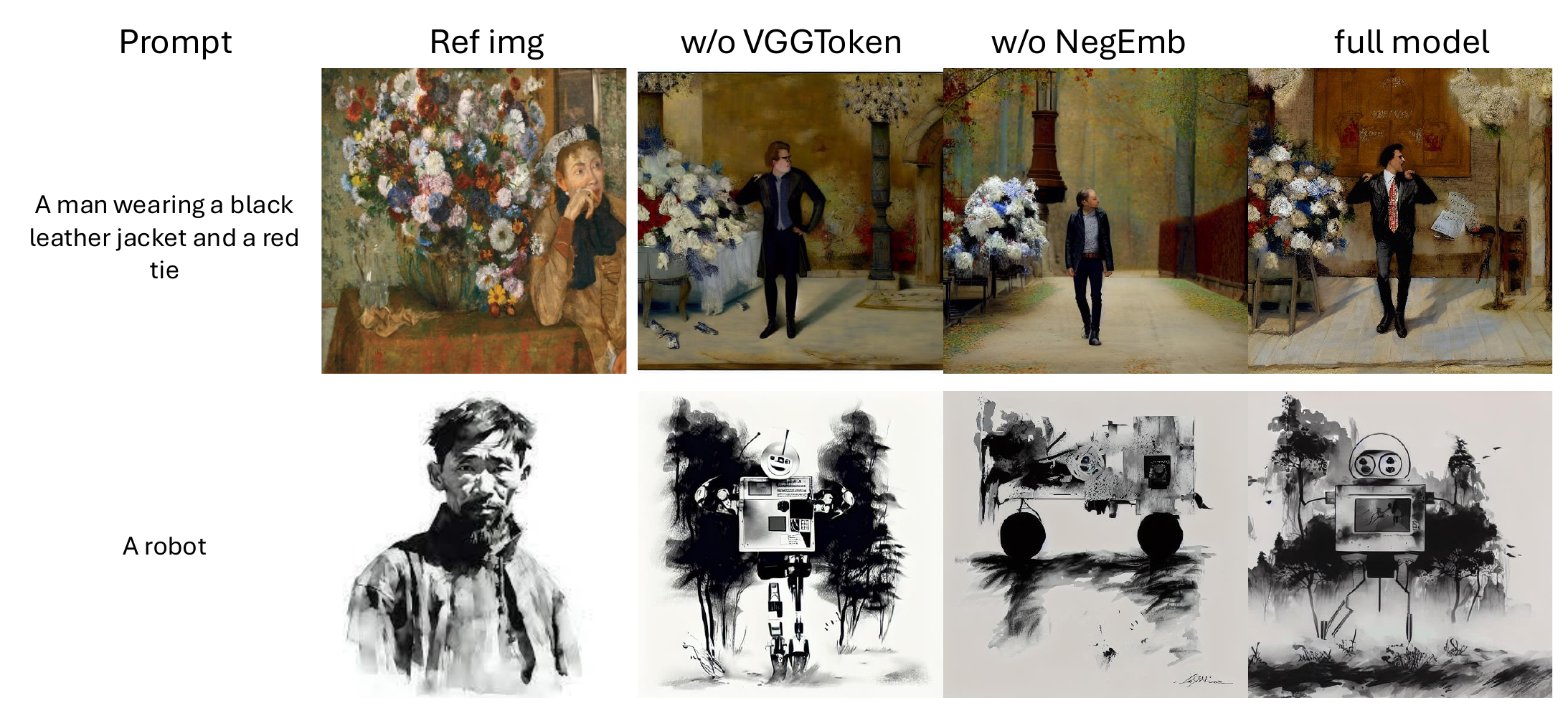}
    \vspace{-0.1in}
    \caption{Images generated by different model variants.}
    \label{fig:ablation1}
    \vspace{-0.05in}
\end{figure}

\noindent\textbf{Design of style embedding module.} We consider two variants together with the full model for the style embedding module: not using the Gram-based descriptor to regulate the attention layers (w/o Gram), and not engaging the semantic descriptor (w/o NegEmb). The results are shown in Fig.~\ref{fig:ablation1}. The style of images generated by model without Gram is generally less mimic. Meanwhile, the model without NegEmb not only has worse style but also suffers from mistaken semantic meaning.

\begin{figure}[htb]
    \centering
    \includegraphics[width=0.95\linewidth]{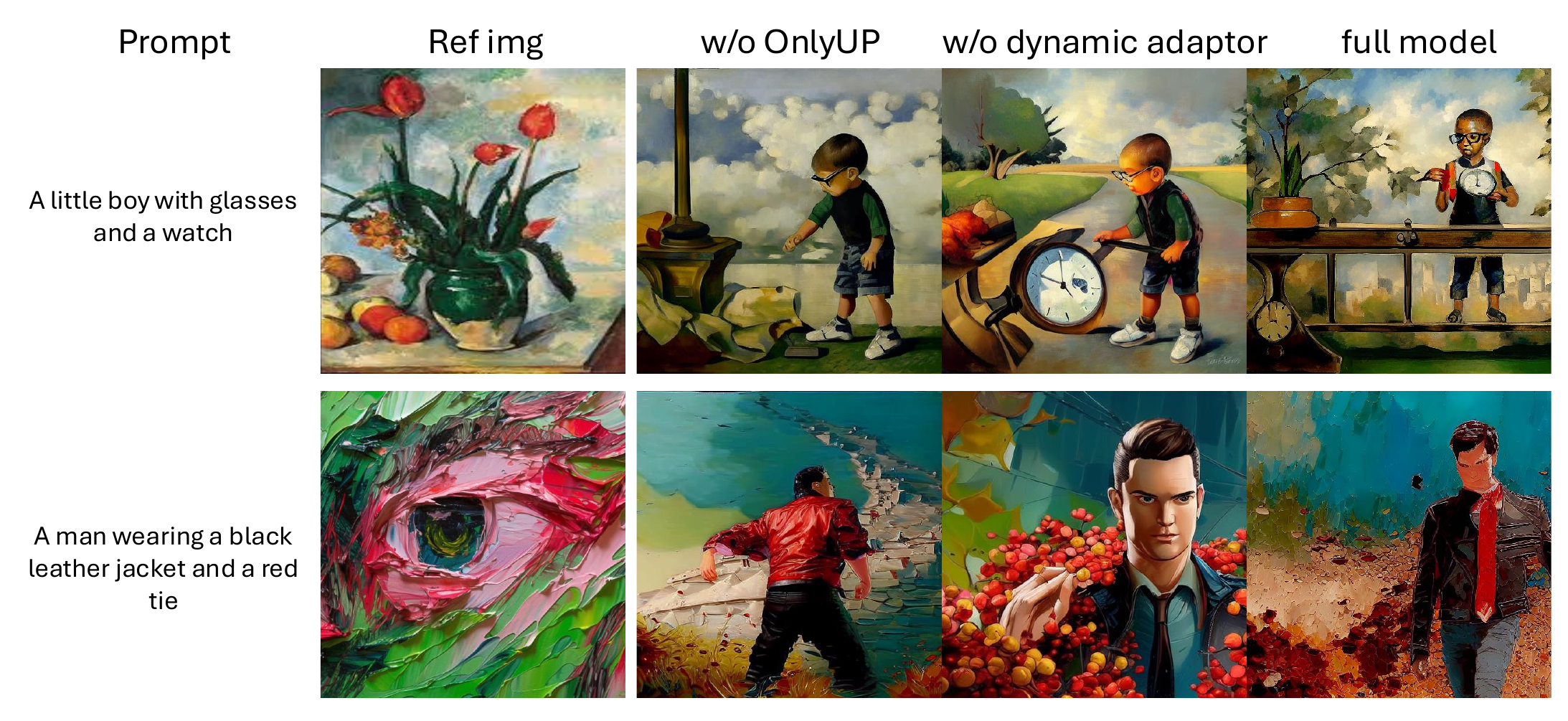}
    \vspace{-0.1in}
    \caption{Images generated by different model variants.}
    \label{fig:ablation2}
    \vspace{-0.05in}
\end{figure}

\noindent\textbf{Design of attention adapter.} We illustrate the role of different parts in the proposed dynamic attention adapter in Fig.~\ref{fig:ablation2}. When the model adopts attention adapter for all UNet attention layers  instead of only the upsample ones, the images generally have problem of mistaken semantic meaning. For example, the clock is missing in the first row, and the cloth color is mistaken in the second row. Also, it is obvious that when only using the same cross-attention adapter as in StyleAdapter, the generated images show inconsistent and undesirable styles, which can be attributed to the limited capacity to such strategy. Interestingly, we find in Tab.~\ref{tab:ablation-quant} that when not using dynamic adapter, the model has a very different tendency compared with the full model, with much better text prompt similarity but much worse style similarity. This is reasonable since the proposed dynamic adapter greatly strengthens the impact of style embedding during both self-attention and cross-attention. In this way, the cross-attention with text prompts is weakened is disguise. In general, adopting dynamic adapter can make a good balance between text prompt fidelity and style fidelity.

\begin{figure}[htb]
    \centering
    \includegraphics[width=\linewidth]{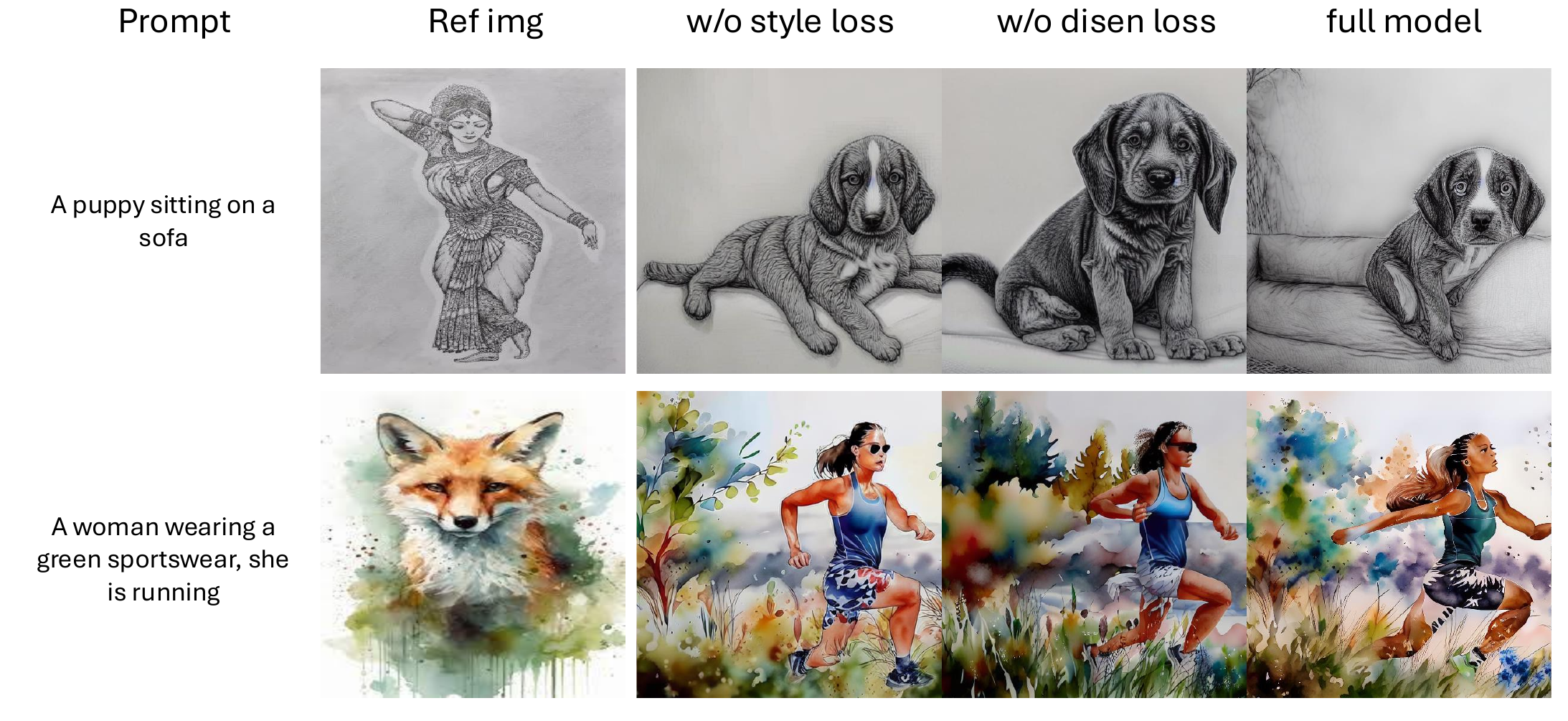}
    \vspace{-0.1in}
    \caption{Images generated by different model variants.}
    \label{fig:ablation3}
    \vspace{-0.05in}
\end{figure}

\noindent\textbf{Effectiveness of different objective functions.} In Fig.~\ref{fig:ablation3} we inspect the efficacy of two objectives introduced in Sec.~\ref{sec:loss}. The results directly support our claim that the gram consistency loss can enhance the style in generated images and the semantic disentangle loss can make the model tell apart semantic and style information from reference images, thus better handling contents in the text prompts. Note that in one-shot setting, using $\mathcal{L}_{style}$ leads to larger improvement in terms of text similarity compared with using $\mathcal{L}_{disen}$. We think this may be because $\mathcal{L}_{style}$ may also help model to extract better style descriptor so that the generated images share more similar styles with the reference ones, thus leading to higher text similarity in this setting.

\section{Conclusion}
We try to solve Stylized Text-to-Image Generation in this paper. A novel model is proposed to solve the problems of misinterpreted style and inconsistent semantics which are suffered by previous methods such as StyleAdapter. The improvement mainly comes from the Mixed Style Descriptor, in which multiple sources and used to achieve comprehensive style embeddings and eliminate the semantic information from style reference images, and the dynamic attention adapter, in which style embeddings dynamically interact with attention layers in diffusion UNet. Extensive experiments have been conducted to show the efficacy of our proposed method as a powerful stylization method, which can be widely applied to real-life scenarios.

\newpage
\section{Experiment details}

\paragraph{Competitors.}  We follow the original training protocol to train StyleAdapter, and utilize 500 and 1000 iterations for LoRA and TI respectively to achieve suitable performance and avoid overfitting. To adapt the style transfer methods to our setting, we first utilize pretrained SD v1.5 to generate an image according to the text prompt, then transfer its style using the corresponding methods. For StyleAlign, we first adopt DDIM-inversion~\cite{mokady2023null} to invert reference images back to noise, and then attach it to other images to be generated.

\begin{figure*}[htp]
    \centering
    \includegraphics[width=\textwidth]{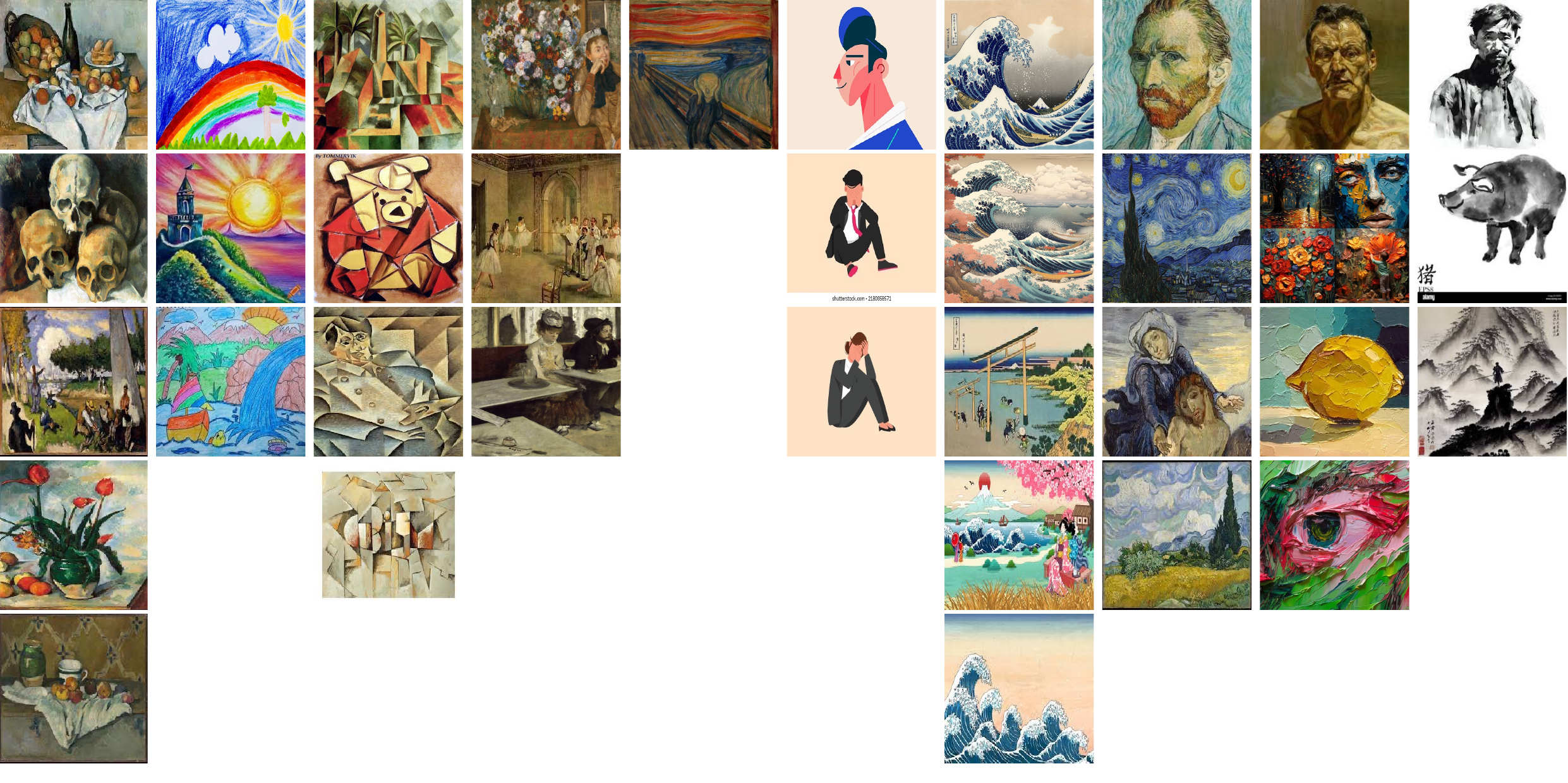}
    \caption{Style reference images used in this paper. Each column denotes a style, in which the images in the first row are used in one-shot experiments, and all images are used in multi-shot experiments. Notations for each column from left to right: \textit{Cezanne, crayon, cubism, Degas, expressionism, flat cartoon, ukiyoe, Van Goah, impasto, ink.}}
    \label{fig:supp-style}
\end{figure*}

\begin{figure*}[htp]
    \centering
    \includegraphics[width=\textwidth]{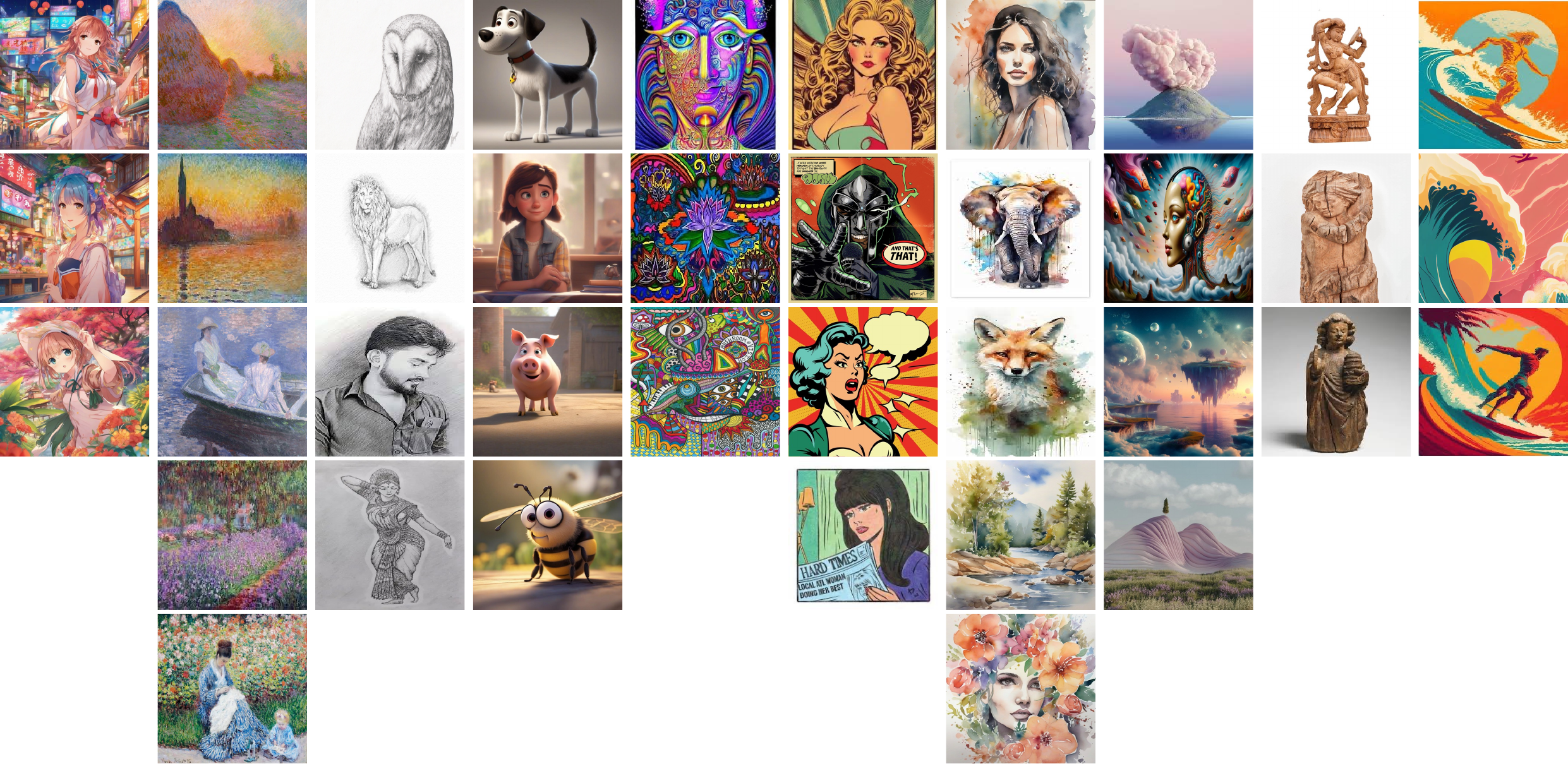}
    \caption{Style reference images used in this paper. Each column denotes a style, in which the images in the first row are used in one-shot experiments, and all images are used in multi-shot experiments. Notations for each column from left to right: \textit{japan anime, Monet, pencil, pixar, psychedelic, america anime, watercolor, surreal, wooden, surf.}}
    \label{fig:supp-style2}
\end{figure*}

\paragraph{Style reference images.} To make sure our experiments is extensive enough to show the generalization ability of our method, we manually design the style reference image set, which are shown in Fig.~\ref{fig:supp-style} and Fig.~\ref{fig:supp-style2}, with the search results provided by Google with key words sets as the notations as depicted in the captions. Our style reference images cover different style concepts such as artistic contents, shapes, colors and textures, which can better support our conclusion that the proposed AnyArt is capable for various cases.

\paragraph{Styles used in figures in main paper.} In order to make some results (Fig.1 and Fig.6) in main paper simple and easier to understand, we omit the detailed style reference images and summarize the used styles here:
\begin{itemize}
    \item Fig.1 from up to bottom: \textit{america cartoon, wooden, cubism, Van Goah}.
    \item Fig.6 from up to bottom: \textit{pencil, watercolor, monet, impasto}.
    \item Fig.7 from up to bottom: \textit{Degas, ink}.
    \item Fig.8 from up to bottom: \textit{Cezanne, impasto}.
    \item Fig.9 from up to bottom: \textit{pencil, watercolor}.
\end{itemize}

\section{More discussion \label{sec:discussion}}
\paragraph{Limitations.} We would like to highlight two limitations of our method. First, the mixed style descriptor (MSD) relies on a patch-level transformer. In this way, it is less efficient for this model to process too many style reference images. While such a scenario is to some extent unrealistic, since it is generally sufficient to represent a specific style with less than 10 images, solving this problem can be related to improving vision transformer structures, which can be taken as future works. Second, the proposed method is only available for style conditions in the form of images. Other forms such as texts, videos and 3D data are not considered in this work and can be solved in the future.

\paragraph{Broader impacts.} Our work will not lead to significant negative social impacts. Problems such as privacy invasion and misinformation can be also attributed to normal image generative models. Solving such problems would be a large future research topic.

\begin{figure}
    \centering
    \includegraphics[width=\linewidth]{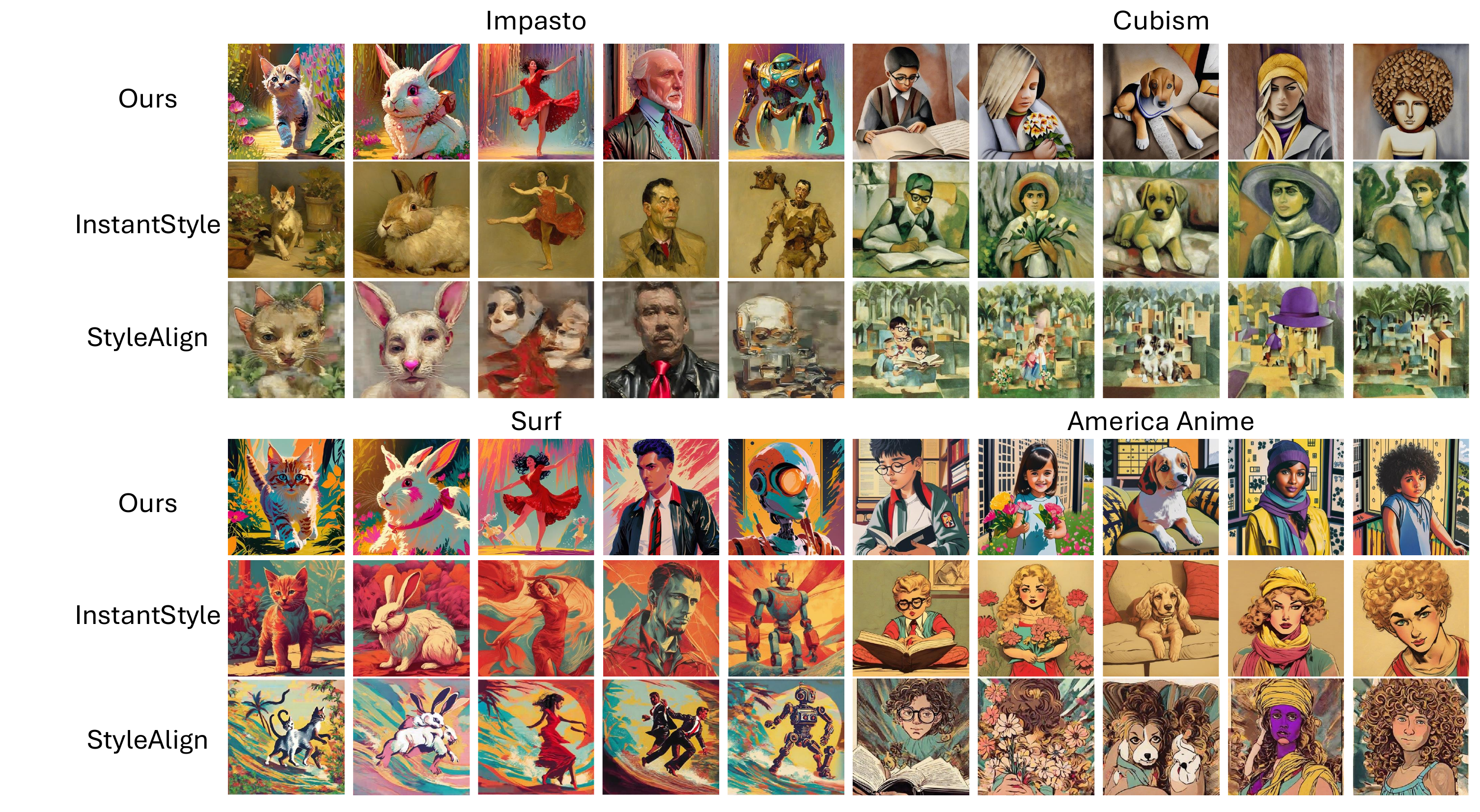}
    \caption{Qualitative comparison with InstantStyle and StyleAlign using SDXL as backbone.}
    \label{fig:instant-style}
\end{figure}

\paragraph{Comparison with InstantStyle-SDXL and StyleAlign.} In Fig.~\ref{fig:instant-style} we present the comparison between our method and InstantStyle and StyleAlign, both using SDXL as backbone network. We find that while InstantStyle does perform better with SDXL than SD1.5, suffering less from content leakage, it tends to generate images with classic art styles such as paintings. This makes the generated results less similar to the reference images. On the other hand, InstantStyle-SDXL still cannot handle styles such as cubism and impasto, which can be well modelled by our method. Moreover, StyleAlign is generally worse than the other two methods.

\begin{figure}
    \centering
    \includegraphics[width=\linewidth]{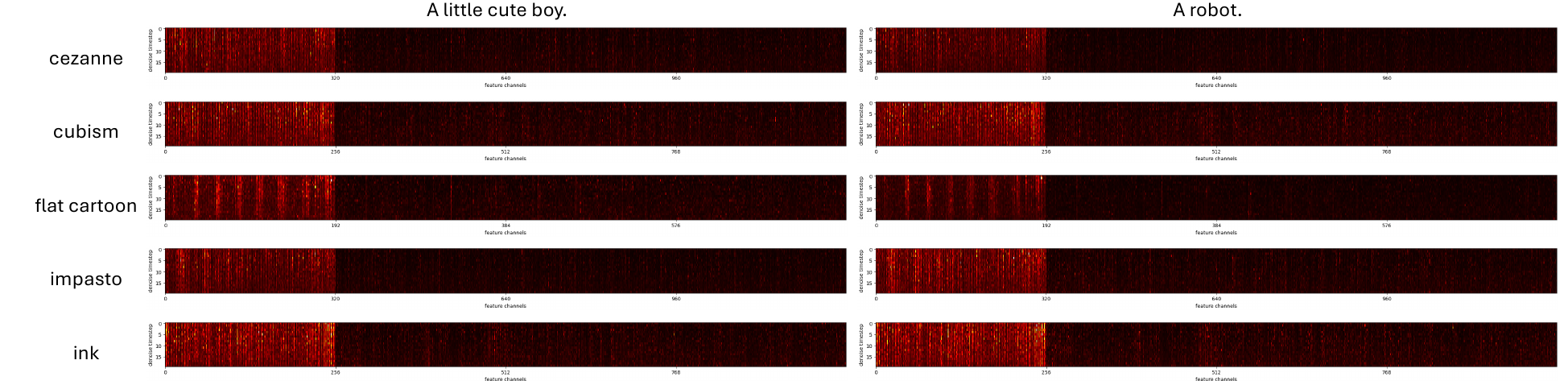}
    \caption{Attention weight visualization between style embedding and frequency domain features of style reference images. Each column represents the same text prompt and each row represents the same style, which are listed in the figure. Zoom in for more details.}
    \label{fig:supp-attn}
\end{figure}

\paragraph{Effectiveness of frequency domain decomposition.} In the mixed style descriptor (MSD), we utilize the discrete wavelet transform to first decompose the patch level features of style reference images into low-frequency and high-frequency features. To see how this process can help our module, we visualize the attention weights for two sets of reference images among different denoising time steps in Fig.~\ref{fig:supp-attn}. Three main phenomenon can be concluded: (1) Style embeddings concentrate more on low frequency features, which is reasonable since low-frequency features contain information such as color. (2) The patterns of feature usage are consistent among two prompts for each style, while being different for different styles. (3) For high-frequency features, different timesteps generally focus on different information. These results can sufficiently support our design of decomposing the image features regarding frequency.

\paragraph{Versatility of our method.} To show that our method is generalizable enough, we further apply our method to pretrained SDXL, which is an advanced version of SD. The results are shown in Fig.~\ref{fig:supp-sdxl}. We can find that basically the proposed method can introduce correct style to pretrained SDXL. Since there is significant gap between the prior knowledge learned by SDXL and SD1.5, the generated images also show different patterns. The results show that SDXL can better handle styles regarding lines and colors, while SD1.5 can provide better global-level styles. Moreover, SDXL can make better balance between style and general image aesthetic. The human faces generated by SDXL are more proper, thanks to its larger capacity.

\begin{figure}
    \centering
    \includegraphics[width=\linewidth]{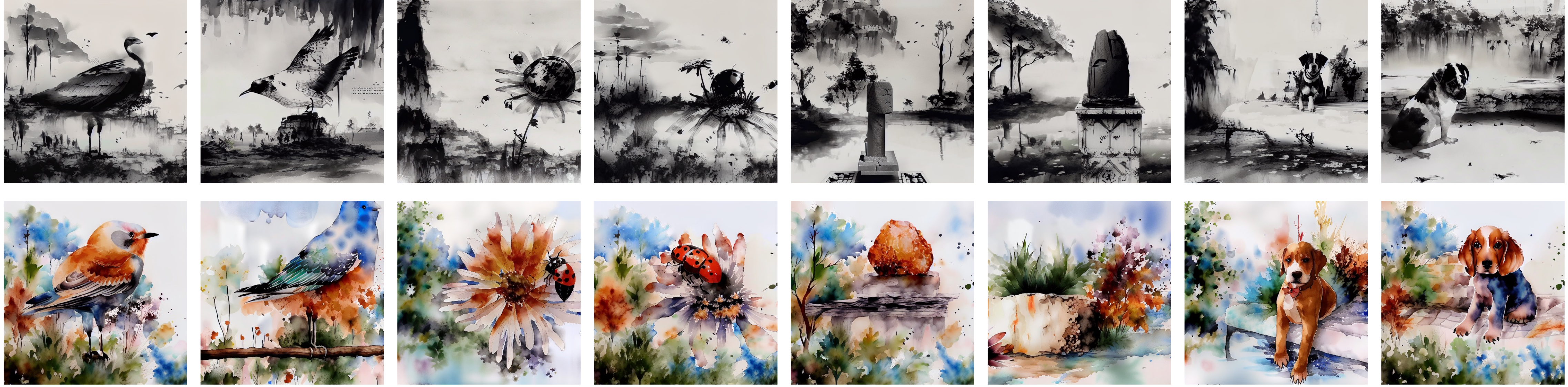}
    \caption{Multi-shot qualitative results with non-related objects added to the caption of style reference images. Prompts for every two columns from left to right: \textit{A bird in a word; A daisy with a ladybug on it; A stone with a crack in it, holding a plant growing out of it; A puppy sitting on a sofa}.}
    \label{fig:multi-shot-nonsense}
\end{figure}

\paragraph{Reasonableness of negative semantic embedding.} One would ask whether it is proper to directly subtract style token part in $\hat{z}_{caption}$ from $\hat{z}_{CLIP}$ and if the subtracted vector could inadvertently contain elements of the negative prompt (e.g., “reading a book”), rather than purely style information. Note that after each subtraction an attention is further applied to $\hat{z}^{CLIP}$, where unrelated negative prompts would be weakened. To verify this we provide several examples on multi-shot ink and watercolor styles. Specifically, we add a non-related caption ‘\textit{There is a robot, a UFO and a monster in the image}.’ to the caption of each reference image. The results are presented in Fig.~\ref{fig:multi-shot-nonsense}, in which the semantics of generated images do not degrade compared with original ArtWeaver.

\begin{figure*}[htb]
    \centering
    \includegraphics[width=\textwidth]{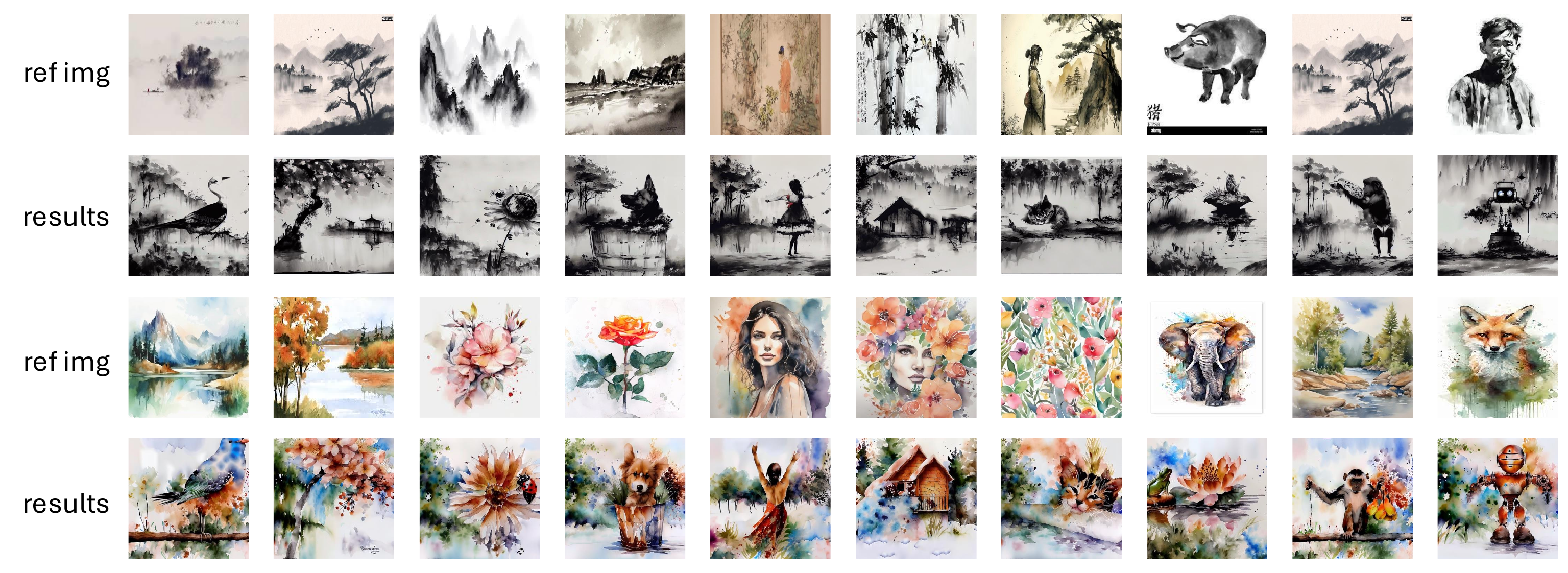}
    \caption{Qualitative results of ArtWeaver under 10-shot setting. The first and third rows contain the reference images, the other rows show the generated results.}
    \label{fig:10-shot}
    \vspace{-0.1in}
\end{figure*}

\begin{figure*}[htb]
    \centering
    \includegraphics[width=\textwidth]{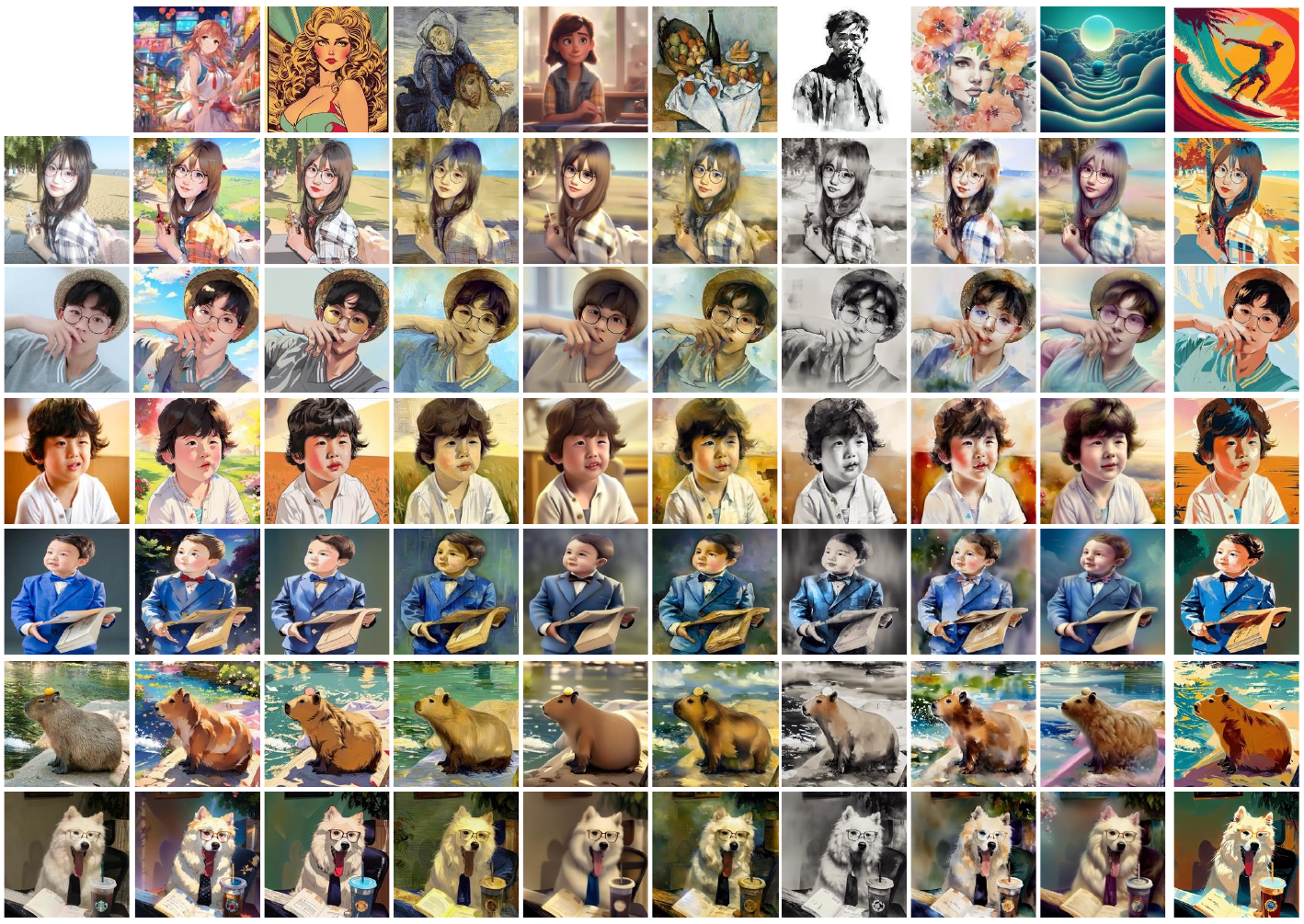}
    \caption{Image to image results generated by our model.}
    \label{fig:img2img}
\end{figure*}

\begin{figure*}
    \centering
    \includegraphics[width=0.8\textwidth]{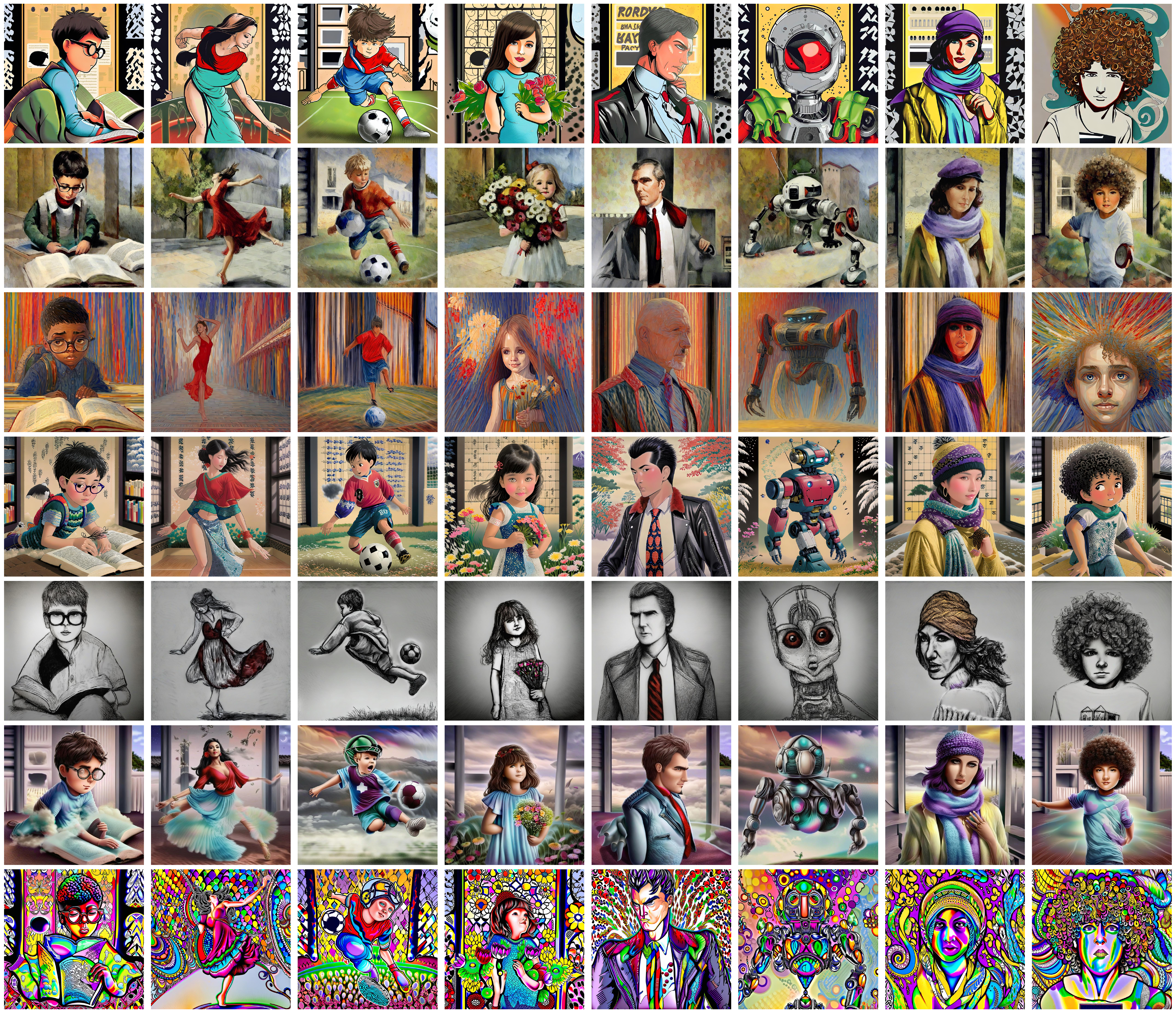}
    \caption{Quantitative results of our proposed method with SDXL in multi-shot setting. Styles used in each row from up to bottom: \textit{america anime, Cezanne, Expressionism, ukiyoe, pencil, surreal, psychedelic}.}
    \label{fig:supp-sdxl}
\end{figure*}

\paragraph{Image-to-image.} Apart from the basic text-to-image task, we also extend our method to image-to-image task, in which we follow the commonly used pipeline to first and then adopt the ControlNet-Canny~\cite{zhang2023contorlnet} together with our proposed method to generate a new image with target style. The results are shown in Fig.~\ref{fig:img2img}. One can find that when given different reference images such as Japan anime, American anime, Von Goah's painting and pixar anime, etc. The human face in the base image can flexibly change according to the style, which illustrates the effectiveness of our method.

\paragraph{Results with more shots.} To further show the versatility of our proposed method, we conduct a 10-shot experiment, of which the results are presented in Fig.~\ref{fig:10-shot}. The results show that our method, when given reasonable reference images, is robust and well performing under 10-shot setting.

% \begin{figure*}
%     \centering
%     \includegraphics[width=0.8\textwidth]{figures/sdxl_1shot.pdf}
%     \caption{Quantitative results of our proposed method with SDXL in one-shot setting. Styles used in each row from up to bottom: \textit{Cezanne, flat cartoon, cubism, pencil}.}
%     \label{fig:supp-sdxl-1shot}
% \end{figure*}

\section{More qualitative results}

\begin{figure*}[htb]
    \centering
    \includegraphics[width=\textwidth]{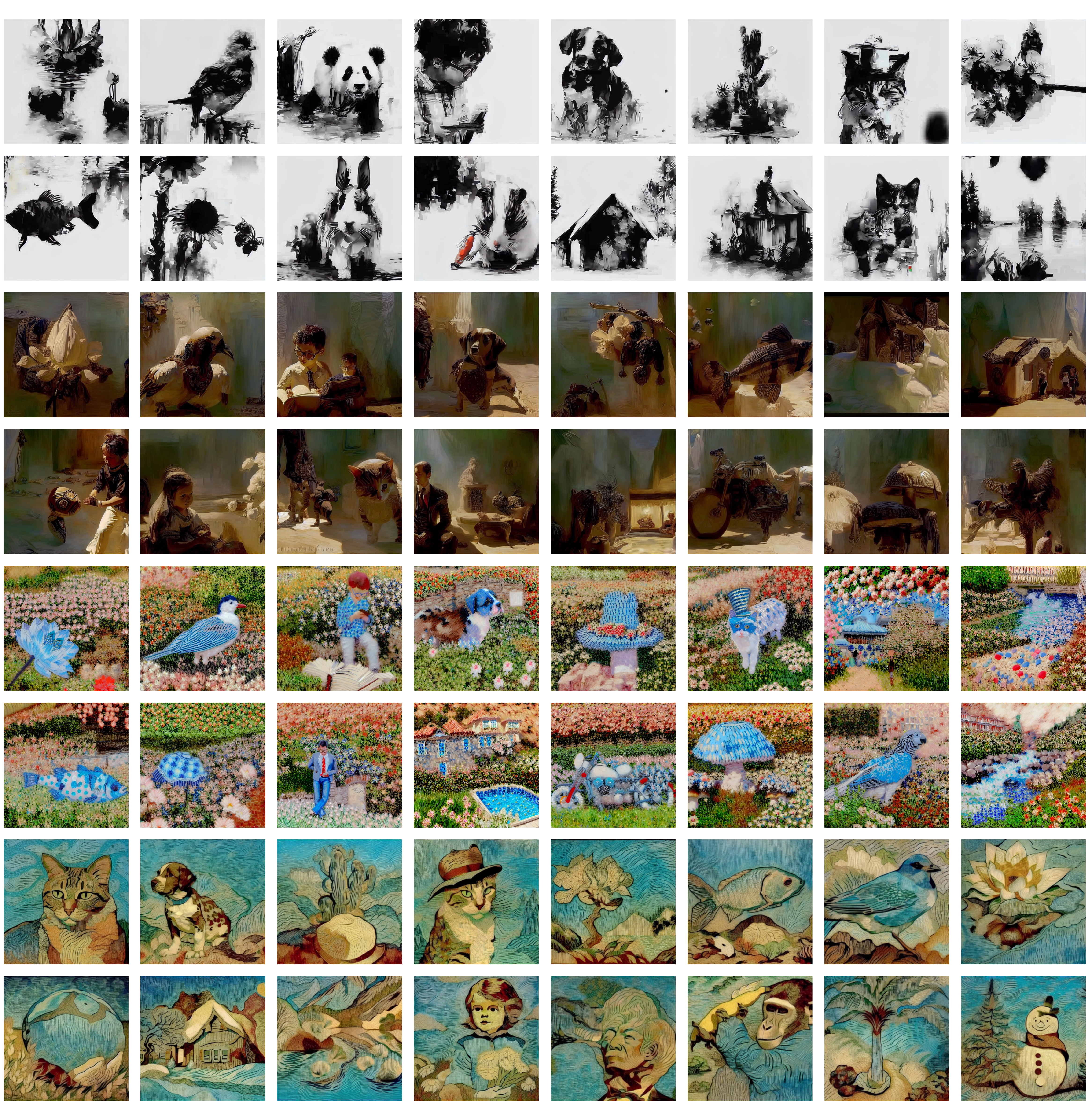}
    \caption{More quantitative results of our proposed method in one-shot setting. Styles used in each two rows from up to bottom: \textit{ink, impasto, Monet, Van Goah}.}
    \label{fig:supp-oneshot}
\end{figure*}

\begin{figure*}[htb]
    \centering
    \includegraphics[width=\textwidth]{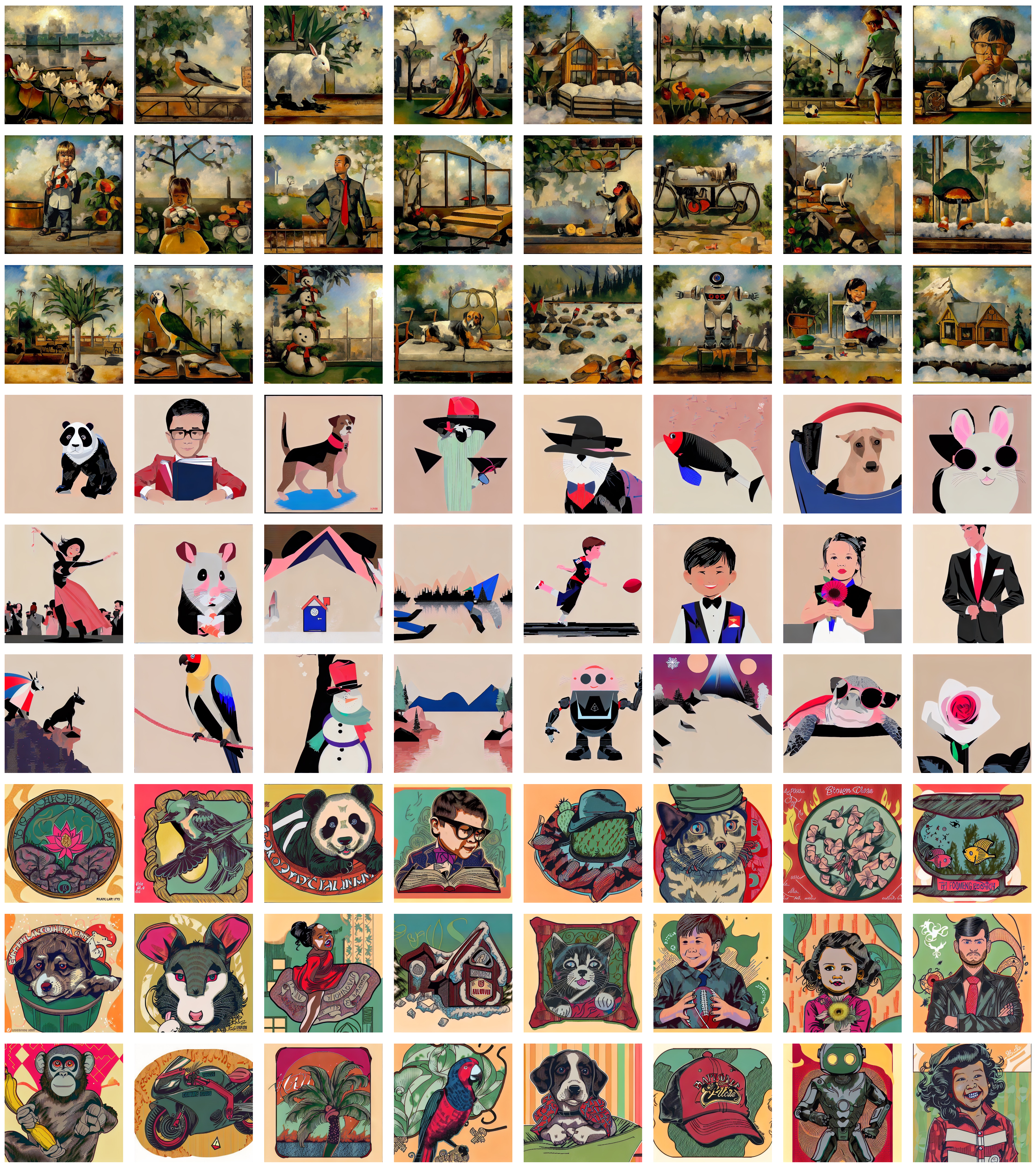}
    \caption{More quantitative results of our proposed method in multi-shot setting. Styles used in each three rows from up to bottom: \textit{Cezanne, flat cartoon, america cartoon}.}
    \label{fig:supp-multishot}
\end{figure*}

% \begin{figure*}[htb]
%     \centering
%     \includegraphics[width=\textwidth]{figures/more_results_multi_shot_2.pdf}
%     \caption{More quantitative results of our proposed method in multi-shot setting. Styles used in each three rows from up to bottom: \textit{surreal, psychedelic, surf}.}
%     \label{fig:supp-multishot2}
% \end{figure*}

We provided more quantitative results in both one-shot and multi-shot settings in Fig.~\ref{fig:supp-oneshot}, Fig.~\ref{fig:supp-multishot}. The images generated with each group of reference images share consistent style, while correctly showing the target objects.

{
    \small
    \bibliographystyle{ieeenat_fullname}
    \bibliography{main}
}
\end{document}